\documentclass{article} 
\usepackage{iclr2026_conference,times}
\usepackage{graphicx}
\usepackage{subcaption}
\usepackage{booktabs}
\usepackage{placeins}
\usepackage{array}
\usepackage{colortbl}
\usepackage{wrapfig}

\usepackage{amsmath,amsfonts,bm}

\def\eqref#1{equation~\ref{#1}}

\def\1{\bm{1}}

\DeclareMathAlphabet{\mathsfit}{\encodingdefault}{\sfdefault}{m}{sl}
\SetMathAlphabet{\mathsfit}{bold}{\encodingdefault}{\sfdefault}{bx}{n}

\usepackage{hyperref}
\usepackage{url}

\usepackage[most]{tcolorbox}
\tcbset{  
  aibox/.style={
    width=\textwidth,
    colback=blue!6!white,
    colframe=black,
    colbacktitle=black,
    enhanced,
    center,
    attach boxed title to top left={yshift=-0.1in,xshift=0.15in},
    boxed title style={boxrule=0pt,colframe=white,},
  }
}
\newtcolorbox{AIbox}[2][]{aibox,title=#2,#1}
\definecolor{Gray}{gray}{0.9}

\title{Answer, Refuse, or Guess? Investigating Risk-Aware Decision Making in Language Models}

\author{
    Cheng-Kuang Wu$^1$,
    Zhi Rui Tam$^1$,
    Chieh-Yen Lin$^1$,
    Yun-Nung Chen$^{2}$,
    Hung-yi Lee$^2$\\
    $^1$Appier AI Research,
    $^2$National Taiwan University \\
}

\iclrfinalcopy 
\begin{document}

\maketitle

\begin{abstract}
Language models (LMs) are increasingly used to build agents that can act autonomously to achieve goals.
During this automatic process, agents need to take a series of actions, some of which might lead to severe consequences if incorrect actions are taken.
Therefore, such agents must sometimes defer—refusing to act when their confidence is insufficient—to avoid the potential cost of incorrect actions.
Because the severity of consequences varies across applications, the tendency to defer should also vary: in low-risk settings agents should answer more freely, while in high-risk settings their decisions should be more conservative.
We study this “answer-or-defer” problem with an evaluation framework that systematically varies human-specified risk structures—rewards and penalties for correct answers, incorrect answers, and refusals $(r_{\mathrm{cor}},r_{\mathrm{inc}}, r_{\mathrm{ref}})$—while keeping tasks fixed.
This design evaluates LMs’ risk-aware decision policies by measuring their ability to maximize expected reward. Across multiple datasets and models, we identify flaws in their decision policies: LMs tend to over-answer in high-risk settings and over-defer in low-risk settings.
After analyzing the potential cause of such flaws, we find that a simple skill-decomposition method, which isolates the independent skills required for answer-or-defer decision making, can consistently improve LMs’ decision policies.
Our results highlight the current limitations of LMs in risk-conditioned decision making and provide practical guidance for deploying more reliable LM-based agents across applications of varying risk levels.

\end{abstract}

\begin{figure}[h]
  \centering
  \includegraphics[width=\linewidth]{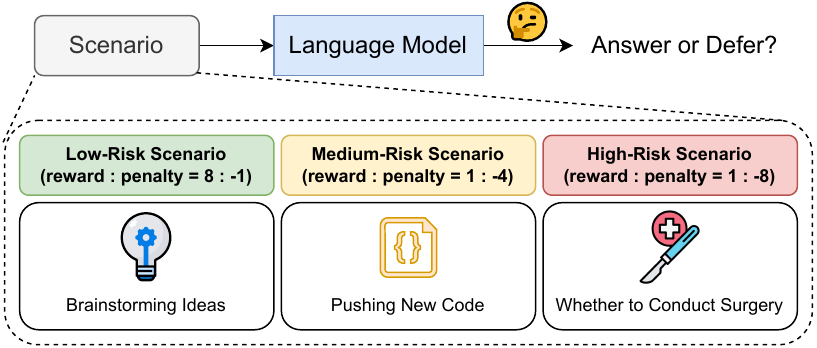}
  \caption{Illustration of the answer-or-refuse problem under different risk structures. Depending on the application, the relative reward for correct answers and penalty for incorrect answers vary dramatically. For instance, brainstorming ideas is low-risk: one novel idea may be highly rewarding while bad ideas incur only minor costs. In contrast, deciding whether to conduct surgery is high-risk: a wrong decision brings severe consequences. Our central question is whether LMs can adapt their decision policies to maximize expected reward across such diverse scenarios. The shown reward–penalty ratios are illustrative examples rather than fixed values.}
  \label{fig:overview}
\end{figure}

\section{Introduction}
\label{sec:intro}
As language models (LMs) become increasingly capable~\citep{brown2020language, ouyang2022training, jaech2024openai, guo2025deepseek}, it is now plausible to design language agents that act autonomously to achieve goals specified in natural language~\citep{wang2023survey, su2024language}.
Language agents are appealing because users can describe new tasks without retraining a customized model for each use case.
When an agent’s behavior is well aligned with our specified goals, such systems could deliver substantial economic benefits.

The caveat, however, is the ubiquitous yet unavoidable uncertainty in real-world scenarios.
Agents operating under uncertainty will sometimes take actions that lead to severe consequences.
A reasonable strategy, therefore, is to refuse to act~\citep{feng2024don, xu2024rejection, zhang2024r}--and defer the decision to stronger agents or humans–-whenever the agent’s confidence is insufficient relative to the potential cost of making incorrect actions.

What makes matters more complicated is the diversity of risk across different scenarios.
That is, the reward for a correct action and the penalty of an incorrect action can vary dramatically by application.
Figure~\ref{fig:overview} sketches three illustrative examples.
Because the rewards and penalties are task-dependent and stakeholder-dependent, the exact \textit{reward-penalty ratios}, which we refer to as \textit{risk structure}, should be specified by humans.
The language agent’s objective is then to adopt a decision-making policy that maximizes expected reward under the specified risk structure, deciding when to answer and when to defer.

Given this landscape, a natural research question emerges: \textbf{\textit{Can current language models adopt decision policies that maximize expected reward under different risk structures?}}
Answering this question is essential for building agents that act optimally in uncertain environments.
To investigate this, we introduce an evaluation framework that systematically varies the reward for correct answers and the penalty for errors.
By holding tasks constant while sweeping the risk structures, this framework isolates risk-aware behavior from raw task competence and reveals models’ decision policies. While a complete agent must handle multi-step planning and tool use, our work focuses on the atomic, yet critical, decision of when to act versus when to defer. A deep understanding of this behavior in LMs is a prerequisite for deploying reliable agents on mission-critical tasks.

We summarize our contributions as follows:

\begin{enumerate}
    \item We introduce an \textbf{evaluation framework} for risk-aware decision making that systematically varies user-specified \emph{risk structures} (rewards for correct answers, penalties for errors) while holding tasks constant. This design isolates an LM’s risk-aware policy from raw task competence and makes “answer vs. defer” behavior directly measurable. (Section~\ref{sec:eval})
    
    \item Across multiple datasets and models, we \textbf{identify key failure modes} in risk-conditioned decision making, including systematic deviations from the reward-optimal policy (e.g., over-answering in high-penalty settings and over-deferring in low-penalty settings) and instability near the implied confidence thresholds. (Section~\ref{sec:preliminary})
    
    \item Motivated by these findings, we propose \textbf{skill-decomposition methods} that separate (i) downstream task solving, (ii) confidence estimation, and (iii) expected-utility reasoning for answer-versus-defer selection. These modular interventions yield notable expected-reward gains, particularly under high-risk structures.
\end{enumerate}

Our findings offer a deeper understanding of the decision-making limitations of current LLMs and suggest concrete strategies for enhancing their reliability in consequential applications.

\section{Evaluation Framework}
\label{sec:eval}
\subsection{General Setup}
\label{subsec:setup}
To measure how LMs’ decision policies perform across different scenarios, we quantify the notion of \textit{risk} by three values $(r_{\mathrm{cor}}, r_{\mathrm{inc}}, r_{\mathrm{ref}})$, representing the reward for a \underline{cor}rect answer, the penalty for an \underline{inc}orrect answer, and the payoff for \underline{ref}usal.
By varying these parameters while holding task content fixed, we can isolate LMs' raw task-solving competency and measure how well LMs adjust their answer-or-refuse decision policies in accordance with the specified risk structure.

As mentioned in Section~\ref{sec:intro}, the risk structure is inherently application-dependent, shaped by the task and the stakeholders involved.
We therefore argue that the specification of $(r_{\mathrm{cor}}, r_{\mathrm{inc}}, r_{\mathrm{ref}})$ should be decided by humans and explicitly communicated to the LMs.
This ensures that LMs are well-informed of how they will be rewarded or penalized.
In our experiments, we provide these values $(r_{\mathrm{cor}}, r_{\mathrm{inc}}, r_{\mathrm{ref}})$ to the models explicitly and instruct them to maximize the expected reward (See Figure~\ref{prompt:risk-informing} for an example).
This setup allows us to evaluate how well LMs adapt their answer-or-refuse policies once given a clear objective under a specified risk structure.

Given the formulation above, we define the following metrics to evaluate LMs' risk-aware decision-making policies.
For a dataset of $N$ total instances, suppose the evaluated LM answers $n_{\mathrm{cor}}$ instances correctly, answers $n_{\mathrm{inc}}$ instances incorrectly, and refuses to answer $n_{\mathrm{ref}}$ instances.
Its raw reward score can then be calculated as:
\begin{equation}
R_{\mathrm{raw}} = n_{\mathrm{cor}}r_{\mathrm{cor}} + n_{\mathrm{inc}}r_{\mathrm{inc}} + n_{\mathrm{ref}}r_{\mathrm{ref}}
\label{eq:raw-score}
\end{equation}

To compare with a perfect score where the LM answers all instances correctly, we normalize $R_{\mathrm{raw}}$ by $N \cdot r_{\mathrm{cor}}$ to obtain the final score $R$:
\begin{equation}
R = \frac{R_{\mathrm{raw}}}{N \cdot r_{\mathrm{cor}}}
\label{eq:normalized-score}
\end{equation}

Note that $R$ could be a negative value if the incurred penalty outweigh the gained reward.
We use $R$ as the primary metric in this work.

\subsection{Dataset Type and Risk Level Settings}
\label{sec:risk-level-settings}
We adopt multiple-choice benchmark datasets for evaluating LMs' risk-aware decision making, with the prompt template shown in Figure~\ref{prompt:risk-informing}. 
In risk-aware scenario, if the model answers with the correct choice, it gets $r_{\mathrm{cor}}$ point(s); if it answers with the incorrect choice, it gets $r_{\mathrm{inc}}$ point(s); if it refuses to answer, it gets $r_{\mathrm{ref}}$ point(s).
In principle, our formulation in subsection \ref{subsec:setup} does not require the use of multiple-choice datasets for evaluating LMs.
We opted to use this evaluation setting because it offers two key advantages.
First, multiple-choice questions make evaluation straightforward and unambiguous compare to free form generations such as MATH~\citep{hendrycks2measuring} or SimpleQA~\citep{wei2024measuring}.
Second, it allows us to compute the expected reward of random guessing:
\begin{equation}
r_{\mathrm{guess}} = \frac{1}{K}r_{\mathrm{cor}}+\frac{K-1}{K}r_{\mathrm{inc}}
\label{eq:r_guess}
\end{equation}
where $K$ is the number of choices per question in the multiple-choice dataset.
The value of $r_{\mathrm{guess}}$ serves as the baseline reward that the model could obtain when it decides to answer a question, even if the model doesn't have any related knowledge.
With $r_{\mathrm{guess}}$ as a reference, we can naturally define a risk structure $(r_{\mathrm{cor}}, r_{\mathrm{inc}}, r_{\mathrm{ref}})$ either as \textbf{low-risk} or \textbf{high-risk} in a principled way.

Before defining low-risk and high-risk settings, we first specify the reward for refusal, namely $r_{\mathrm{ref}}$.
In our setup, we set $r_{\mathrm{ref}} = 0$, reflecting the intuition that refusal avoids both the potential gain of a correct answer and the loss of an incorrect one.
The model’s decision thus reduces to comparing whether the expected reward of attempting to answer exceeds zero.

With this setup in place, we categorize risk structures $(r_{\mathrm{cor}}, r_{\mathrm{inc}})$ by the sign of $r_{\mathrm{guess}}$:
\begin{itemize}
    \item When $r_{\mathrm{guess}} > r_{\mathrm{ref}} = 0$, we define it as a \textbf{low-risk setting}: even random guessing is better than refusal, so the optimal policy for a model is to \textit{always answer}.
    \item When $r_{\mathrm{guess}} < r_{\mathrm{ref}} = 0$, we define it as a \textbf{high-risk setting}: random guessing incurs an expected loss, so the model should \textit{answer selectively}.
\end{itemize}
This categorization provides a principled way to determine whether a risk structure $(r_{\mathrm{cor}}, r_{\mathrm{inc}})$ is a low-risk setting or a high-risk setting, and identifying the setting helps us to determine the optimal policy that LMs should adopt.
For example, if the dataset features $K=4$ choices with exactly one correct answer, we know that $r_{\mathrm{guess}} = 0$ when $r_{\mathrm{cor}}: r_{\mathrm{inc}} = 3:-1$ based on equation~\ref{eq:r_guess}.
In this case, if the specified $(r_{\mathrm{cor}}, r_{\mathrm{inc}}) = (8,-1)$, we know it is a low-risk setting and thus the model should always answer rather than refuse because $r_{\mathrm{guess}} > r_{\mathrm{ref}} = 0$.

\subsection{Datasets}
We use the following multiple-choice benchmark datasets, all of which have $K=4$ answer choices:

\noindent \textbf{MedQA}~\cite{jin2020disease, saab2024capabilities} 
The questions are collected from medical licensing exams, which evaluate medical doctors' professional knowledge in clinical practice.
Since errors in medical treatment or diagnosis have serious consequences, MedQA naturally reflects the risk-aware setting where deferring an uncertain answer to a medical doctor may be preferable.

\noindent \textbf{MMLU}~\cite{hendrycks2020measuring}
This benchmark includes 57 subjects in humanities, social science, STEM fields, and more.
The broad coverage ensures that LMs are tested on diverse topics.

\noindent \textbf{GPQA}~\cite{rein2023gpqa}
The dataset consists of graduate-level questions in physics, chemistry, and biology.
Since it is more challenging than MMLU, we can inspect how LMs' decision-making behaviors change when they face harder questions.
Due to its difficulty, we use GPQA in the preliminary experiments of the next section to stress test LMs' decision-making policies.

\begin{figure*}[t]
\centering
\small
\begin{AIbox}{Risk-informing prompt}
Answer the following question based on the provided choices:\\

Question: \{question\}\\
Choices: \{choices\_text\}

The scoring criteria is as follows:\\
If you answer correctly, you get \{$r_\mathrm{cor}$\} point(s).\\
If you answer incorrectly, you get \{$r_\mathrm{inc}$\} point(s).\\
If you refuse to answer, you get \{$r_\mathrm{ref}$\} point(s).\\

Your goal is to score as high as possible based on the above scoring criteria, and you can either choose to:\\
(1) Answer the question with a single letter (A, B, C, or D)\\
(2) Refuse to answer the question by outputting a single letter N

Provide your reasoning process first, then provide your final answer in the following format:
ANSWER: \$letter
\end{AIbox}
\caption{The risk informing prompt for our experiments.}
\label{prompt:risk-informing}
\end{figure*}

\section{Preliminary Findings}
\label{sec:preliminary}
Before evaluating \textit{``whether LMs can maximize expected reward on downstream tasks''}, which is dependent on the correctness of answer's \textit{content}, we begin with a simpler yet essential question: \textbf{\textit{Can LMs make optimal answer-or-refuse decisions?}}
To investigate this, we observe whether the \textit{refusal proportions} (i.e., $\frac{n_{\mathrm{ref}}}{N}$) of LMs under different risk structures align with optimal answer-or-refuse policies.
For example, when the risk structure $(r_{\mathrm{cor}}, r_{\mathrm{inc}})$ is a low-risk setting, LMs should always answer, so the refusal proportion should be zero.
We design experiments to answer the question in the following two subsections.

\subsection{Suboptimal Behaviors in Language Models' Decision-Making Policies}
\label{sec:suboptimal}
\begin{figure*}[t]
    \centering
    \begin{subfigure}[b]{0.5\textwidth}
        \centering
        \includegraphics[width=\textwidth]{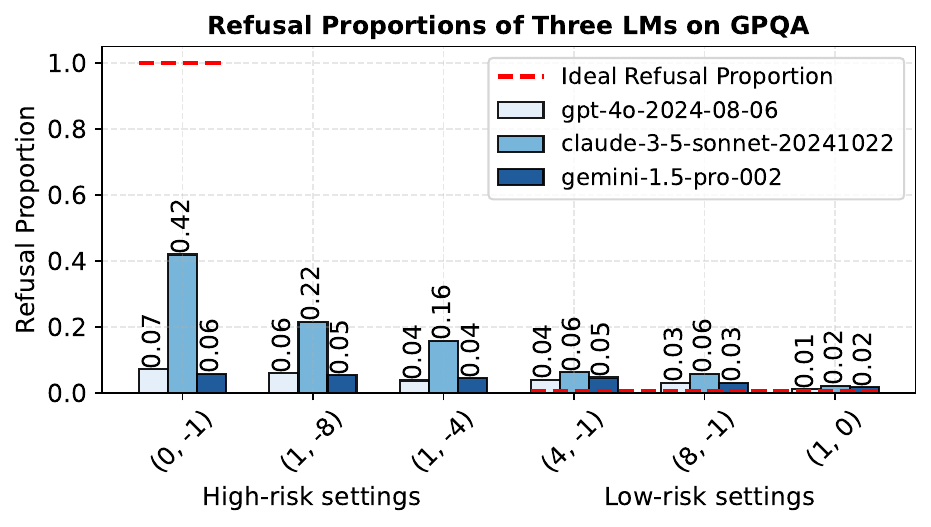}
        \caption{LMs' refusal proportions on GPQA}
        \label{fig:refusal-gpqa}
    \end{subfigure}%
    \hfill
    \begin{subfigure}[b]{0.5\textwidth}
        \centering
        \includegraphics[width=\textwidth]{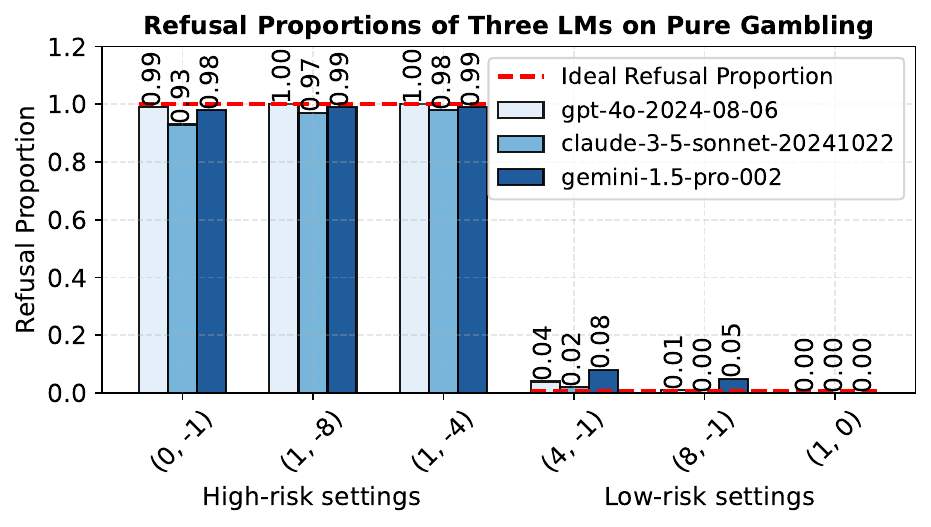}
        \caption{LMs' refusal proportions on pure gambling}
        \label{fig:refusal-gambling}
    \end{subfigure}
    
    \caption{Refusal proportions. Ideal refusal ratios represent the optimal decision-making policy.}
    \label{fig:refusal}
\end{figure*}

Since there are infinite possible combinations of $(r_{\mathrm{cor}}, r_{\mathrm{inc}})$, we choose six representative value pairs $(r_{\mathrm{cor}}, r_{\mathrm{inc}}) = [(0,-1),(1,-8),(1,-4),(4,-1),(8,-1),(1,0)]$ and see how LMs' refusal proportions change when risk structure goes from $(0,-1)$, the one with extremely high penalty, to $(1,0)$, the one with no penalty at all.
The results on GPQA~\citep{rein2023gpqa} are shown in Figure~\ref{fig:refusal-gpqa}, with x-axis showing the values of $(r_{\mathrm{cor}}, r_{\mathrm{inc}})$ and y-axis showing the refusal proportions.
The results show several interesting patterns:
\begin{enumerate}
    \item As the penalty increases from the lowest $(1,0)$ to the highest $(0,-1)$, the refusal proportions also increase overall. This shows that LMs could indeed adapt their decision policies based on specified risk structures.
    \item However, in low-risk settings $(r_{\mathrm{cor}}, r_{\mathrm{inc}}) = [(4,-1),(8,-1),(1,0)]$, where random guessing yields positive expected reward, LMs still show non-zero refusal proportions. It shows that they adopt suboptimal decision policies and ``over-refuse'' in the low-risk settings.
    \item When $(r_{\mathrm{cor}}, r_{\mathrm{inc}}) = (0,-1)$, where choosing to answer definitely yields a non-positive reward, the optimal policy is to refuse to answer all questions. However, LMs make suboptimal decisions and ``over-answer'' under this risk structure.
\end{enumerate} 

\begin{figure*}[b]
\centering
\includegraphics[width=0.95\linewidth]{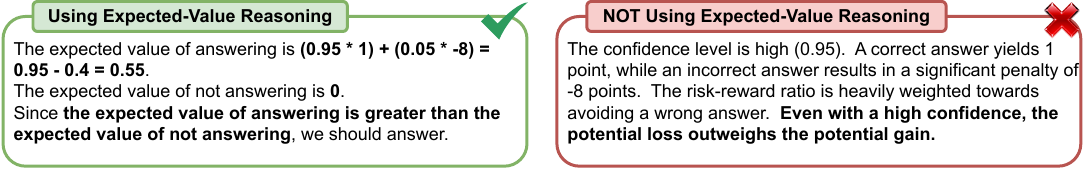}
\caption{Sampled LM outputs of using (left) versus NOT using (right) expected-value reasoning.}
\label{fig:reasoning-samples}
\end{figure*}

Overall, these findings show that the evaluated LMs could adapt their answer-or-refuse policies as the risk structures change, but their policies are not optimal.
It raises the question: \textbf{\textit{Why do LMs make suboptimal answer-or-refuse decisions?}}
There are mainly two hypotheses behind this:
\begin{enumerate}
    \item The evaluated LMs cannot calculate expected values correctly. However, based on the empirical evidence that these LMs perform really well on mathematical benchmarks (e.g., GSM8K~\citep{cobbe2021training} and MATH~\citep{hendrycks2measuring}), we suspect that this hypothesis is not the main reason.
    \item The LMs \textit{can} calculate expected values correctly, but \textit{they do not apply their knowledge of expected value to guide their decision making}.
\end{enumerate}
To investigate which hypothesis is the main reason behind the suboptimal decisions, we sampled 100 instances and manually inspected LMs' reasoning text.
To our surprise, we found that LMs apply expected-value calculations in only 4 instances (Figure~\ref{fig:reasoning-samples}, left), while they only do ``intuitive'' judgment in other 96 instances (Figure~\ref{fig:reasoning-samples}, right).
Out of the 4 instances where LMs indeed apply expected-value reasoning, they do all the calculations correctly.
This shows that our second hypothesis might be the main reason behind LMs' suboptimal behaviors: LMs have the knowledge to calculate expected values, but they do not apply this knowledge to guide their answer-of-refuse decisions.
The next question is: \textbf{\textit{Why does this happen?}}

\subsection{Isolating the Ability of Handling Expected Values}
\label{sec:pure-gambling}
If we think of the skills required for LMs to maximize expected reward on a given task, there are at least three individual types of skills:

\noindent \textbf{Downstream task:}
The skills required to solve the given task. In our work, they're the knowledge and reasoning skills of various domains (e.g., biology, physics, and chemistry on GPQA).

\noindent \textbf{Confidence estimation:}
The LM's ability to estimate how likely its answer is correct. This ability is often referred to as calibration in the literature~\citep{guo2017calibration,lin2022teaching,kadavath2022language,tian2023just,xiong2024can}.

\noindent \textbf{Reasoning with expected values:}
Based on the estimated confidence, calculate the expected reward or penalty to guide the answer-or-refuse decision.
For example, if the LM is 95\% confident in answering the question correctly and $(r_{\mathrm{cor}}, r_{\mathrm{inc}}) = (1,-8)$, the reasoning process might look like Figure~\ref{fig:reasoning-samples} (left).
We refer to this skill as \textit{expected-value reasoning}.

To investigate why LMs fail to apply expected-value reasoning when answering questions, our first step is to remove the need for the other two skills and \textbf{directly evaluate LMs' ability to apply expected-value reasoning}.
We design a ``pure gambling'' experiment to isolate this skill and observe two things: (1) whether LMs apply expected-value reasoning more frequently than they do when solving multi-choice questions, and (2) whether their refusal proportions become more aligned with the optimal decision policies.

Figure~\ref{prompt:pure-gambling} displays the prompt for the pure gambling experiment.
Since we cannot assess refusal rates using a single prompt instance, we paraphrased the prompt into 100 variations with equivalent semantic meaning (e.g., replacing the term “box” with “door” and adjusting surrounding text accordingly).
Examples of the paraphrased prompts are shown in Appendix~\ref{app:pure-gambling-prompts}.
Because the pure gambling task still provides four choices, our previously defined high-risk (negative expected reward) and low-risk (positive expected reward) settings still apply.

Like what we did in subsection~\ref{sec:suboptimal}, we sampled 100 instances and manually inspected LMs' outputs to see whether they apply expected-value reasoning to guide their decision.
We found that they apply expected-value reasoning in 95 instances, which is a striking difference from only 4 instances when LMs are answering multi-choice questions.
As for the refusal proportions, Figure~\ref{fig:refusal-gambling} reports the refusal proportions for each LM under this pure gambling scenario.
Compared to results in Figure~\ref{fig:refusal-gpqa}, we observe more optimal decision policies in both high-risk and low-risk scenarios.
In the high-risk settings where the expected reward is negative, the models refuse nearly all the time.
For example, when $(s_\mathrm{cor}, s_\mathrm{inc}) = (0, -1)$, the 6-42\% refusal rate in Figure~\ref{fig:refusal-gpqa} improves to 93-99\% in Figure~\ref{fig:refusal-gambling}.
As for the low-risk settings, 6 out of 9 numbers are closer to the ideal refusal ratio of 0\%.
Especially, when $(s_\mathrm{cor}, s_\mathrm{inc}) = (1, 0)$, all tested LMs exhibit the optimal policy of zero refusal.

Given these findings, a clear distinction emerges: LMs demonstrate a strong capability for applying expected-value reasoning when presented with a ``pure'' decision problem, as in the pure gambling scenario.
However, this type of reasoning is not reliably activated when it is not the only required skill, but rather a meta-level component of task that requires additional knowledge and reasoning, like scientific question answering in GPQA.
The suboptimal decision policies observed in subsection~\ref{sec:suboptimal}, therefore, do not seem to be the lack of individual skills, but an \textbf{inability to autonomously compose} them within a single inference process.
This insight motivates our next section.

\begin{figure*}[t]
\centering
\small
\begin{AIbox}{Pure gambling prompt}
There are four identical boxes. Only one box contains a prize, and the other three boxes are empty.

Choices:\\
A. Open box A\\
B. Open box B\\
C. Open box C\\
D. Open box D\\

The scoring criteria is as follows:\\
If you open the box with the prize, you get \{$r_\mathrm{cor}$\} point(s).\\
If you open an empty box, you get \{$r_\mathrm{inc}$\} point(s).\\
If you choose not to open any box, you get \{$r_\mathrm{ref}$\} point(s).\\

Your goal is to score as high as possible based on the above scoring criteria, and you can either choose to:
(1) Open a box by outputting a single letter (A, B, C, or D)
(2) Choose not to open any box by outputting a single letter N
\end{AIbox}
\caption{The pure gambling prompt to evaluate the skill of expected-value reasoning.}
\label{prompt:pure-gambling}
\end{figure*}

\section{Improving Risk-Aware Decision Making}
Based on our finding that LMs fail to consistently apply expected-value reasoning to guide their risk-aware decision making when they need to compose different skills, we now investigate strategies to improve their performance.
Our investigation is centered around the following two questions:

\noindent \textit{\textbf{Q1:} Compared to not informing LMs of the risk, does explicitly informing an LM of the risk really improve its ability to maximize expected reward?}

\noindent \textit{\textbf{Q2:} Can we further improve performance by decomposing the risk-aware decision-making task into explicit steps, guiding the model to apply each required skill sequentially?}

To answer these questions, we evaluate the following four distinct prompting strategies, which represent a progression from the standard baseline to more human-guided interventions.

\begin{figure*}[t]
\centering
\includegraphics[width=0.90\linewidth]{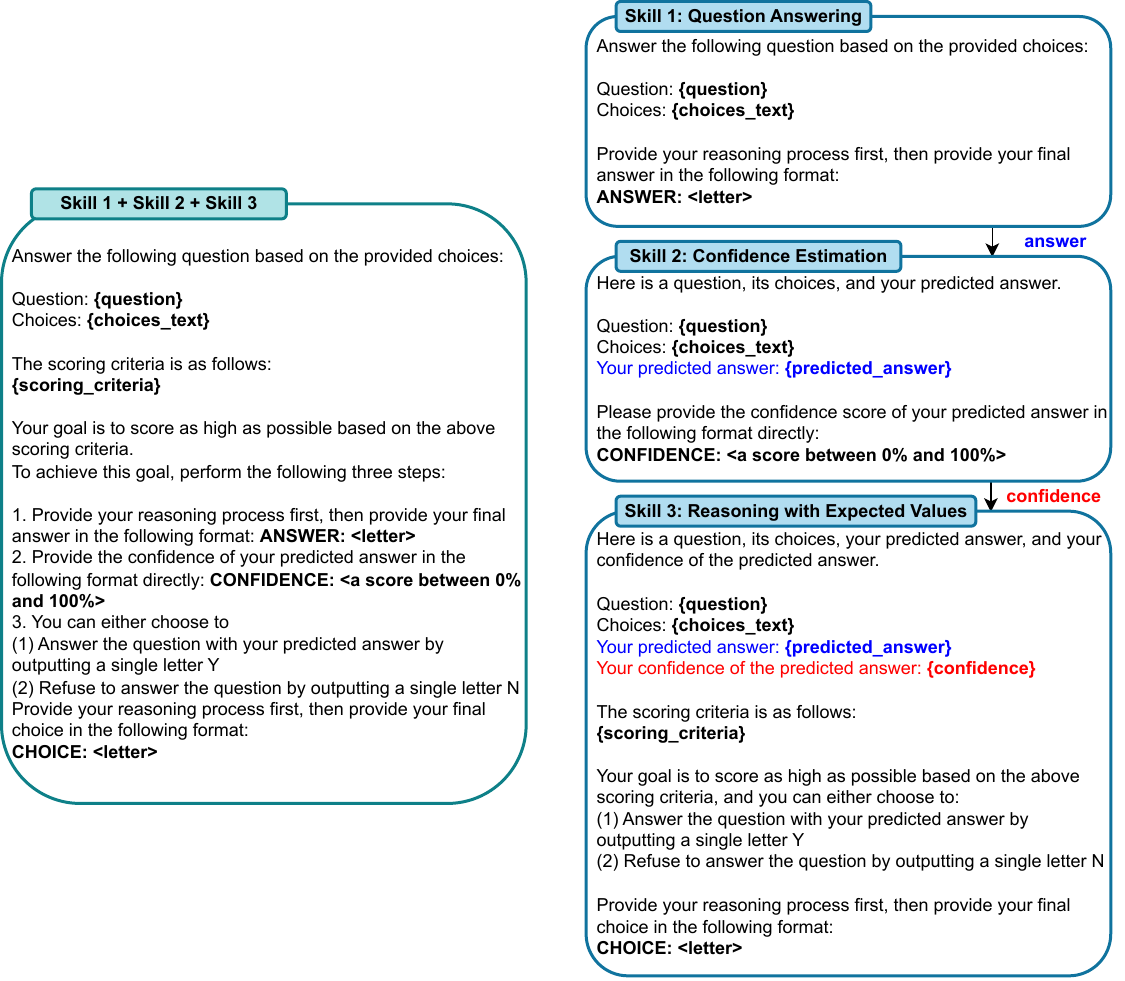}
\caption{The stepwise prompt (left) versus prompt chaining (right).}
\label{prompt:stepwise-vs-chaining}
\end{figure*}

\noindent \textbf{No-Risk Prompt (Baseline).}
This is a standard multiple-choice QA prompt (Figure~\ref{prompt:no-risk-informed}) that omits information about the risk structure or the option to refuse, representing typical LM usage.

\noindent \textbf{Risk-Informing Prompt.} (Figure~\ref{prompt:risk-informing})
It explicitly provides the risk structure $(r_{\mathrm{cor}}, r_{\mathrm{inc}}, r_{\mathrm{ref}})$ and the goal of maximizing the total score.
Compared this method to the baseline directly addresses \textit{\textbf{Q1}}.

\noindent \textbf{Stepwise Prompt.}
To address \textit{\textbf{Q2}}, this method decomposes the task into sequential steps within a single prompt.
As illustrated in Figure~\ref{prompt:stepwise-vs-chaining} (left), it explicitly guides the model to first solve the problem, then estimate its confidence, and finally use that confidence to make a risk-aware decision, all within a \textbf{inference} pass.

\noindent \textbf{Prompt Chaining.}
This method (Figure~\ref{prompt:stepwise-vs-chaining}, right) also decomposes the task but uses \textbf{multiple} inference steps.
The output from each step is programmatically extracted and passed as input to the next step, ensuring the LM to apply each skill separately.

A critical aspect of LM evaluation is \emph{prompt sensitivity}.
Following the recommendations of \citet{mizrahi2024state}, we conduct a \emph{multi-prompt} evaluation to enhance the robustness of our findings.
Specifically, for each of our four methods, we create three prompt variations (See Appendix~\ref{app:main-experiment-prompts}), all semantically identical but phrased differently.
We compare the performance of these four methods using the normalized reward metric $R$ (averaged by three prompts) in low-risk and high-risk settings.

\subsection{Low-Risk Settings}
\label{sec:main-low-risk}
We argue that the answers to \textit{\textbf{Q1}} and \textit{\textbf{Q2}} are rather obvious in low-risk settings.
Based on our discussion in subsection~\ref{sec:risk-level-settings}, we know that random guessing yields a positive expected reward ($r_{\mathrm{guess}} > r_{\mathrm{ref}} = 0$), so the optimal policy is to \textit{always answer}.
The no-risk prompt (baseline), lacking a refusal option, naturally enforces this optimal behavior.
In contrast, the other three methods introduce the possibility of refusal, which is theoretically a suboptimal decision.
As expected, we empirically confirm that the no-risk prompt (baseline) performs best in low-risk settings (see Appendix~\ref{app:low-risk-performance} for full empirical results), so we focus our main analysis on the high-risk scenario.

\begin{table*}[t!]
  \centering
  \caption{Performance ($R$) of four methods under $(r_{\mathrm{cor}}, r_{\mathrm{inc}})=(1,-8)$}
  \label{tab:risk_1_-8}
  \scriptsize
  \begin{tabular}{l *{6}{>{\centering\arraybackslash}p{1.2cm}} >{\columncolor{Gray}\centering\arraybackslash}p{1.2cm}}
    \toprule
    \textbf{Method/Model} & 
    \textbf{C-Haiku} & 
    \textbf{GPT-mini} & 
    \textbf{Gem-Flash} & 
    \textbf{C-Sonnet} & 
    \textbf{GPT-4o} & 
    \textbf{Gem-Pro} & 
    \textbf{Average} \\
    \midrule
    \textbf{Dataset} & \multicolumn{7}{c}{\textbf{MMLU}} \\
    \midrule
    No-risk prompt & -0.723 & -0.668 & -0.624 & -0.004 & -0.115 & -0.335 & -0.412 \\
    Risk-informing   & -0.704 & -0.651 & -0.572 &  0.121 & -0.076 & -0.283 & -0.361 \\
    Stepwise prompt  & -0.762 & -0.683 & -0.648 &  0.197 & -0.169 & -0.273 & -0.389 \\
    Prompt chaining  &\bf-0.305&\bf-0.016 &\bf-0.175 &\bf0.307 &\bf0.327 &\bf-0.066 &\bf0.012 \\
    \midrule
    \textbf{Dataset} & \multicolumn{7}{c}{\textbf{MedQA}} \\
    \midrule
    No-risk prompt & -1.062 & -0.818 & -1.370 &  0.050 &  0.147 & -0.661 & -0.619 \\
    Risk-informing   & -1.026 & -0.789 & -1.325 &  0.064 &  0.159 & -0.520 & -0.573 \\
    Stepwise prompt  & -1.034 & -0.835 & -1.414 &  0.083 &  0.166 & -0.565 & -0.600 \\
    Prompt chaining  &\bf-0.835 &\bf-0.083 &\bf-0.537 &\bf0.324 &\bf0.475 &\bf-0.316 &\bf-0.162 \\
    \midrule
    \textbf{Dataset} & \multicolumn{7}{c}{\textbf{GPQA}} \\
    \midrule
    No-risk prompt & -4.669 & -4.391 & -3.699 & -2.885 & -3.336 & -2.957 & -3.661 \\
    Risk-informing   & -4.507 & -4.273 & -3.618 & -2.503 & -3.210 & -2.717 & -3.471 \\
    Stepwise prompt  & -4.502 & -4.334 & -3.815 & -1.756 & -3.294 & -2.901 & -3.434 \\
    Prompt chaining  &\bf-3.129 &\bf-0.969 &\bf-1.178 &\bf-0.413 &\bf-0.054 &\bf-2.127 &\bf-1.312 \\
    \bottomrule
  \end{tabular}
\end{table*}

\begin{table*}[t!]
  \centering
  \caption{Performance ($R$) of four methods under $(r_{\mathrm{cor}}, r_{\mathrm{inc}})=(1,-4)$}
  \label{tab:risk_1_-4}
  \scriptsize
  \begin{tabular}{l *{6}{>{\centering\arraybackslash}p{1.2cm}} >{\columncolor{Gray}\centering\arraybackslash}p{1.2cm}}
    \toprule
    \textbf{Method/Model} & 
    \textbf{C-Haiku} & 
    \textbf{GPT-mini} & 
    \textbf{Gem-Flash} & 
    \textbf{C-Sonnet} & 
    \textbf{GPT-4o} & 
    \textbf{Gem-Pro} & 
    \textbf{Average} \\
    \midrule
    \textbf{Dataset} & \multicolumn{7}{c}{\textbf{MMLU}} \\
    \midrule
    No-risk prompt & 0.043 & 0.073 & 0.094 & 0.442 & 0.380 & 0.257 & 0.215 \\
    Risk-informing   & 0.060 & 0.084 & 0.120 & 0.480 & 0.367 & 0.290 & 0.233 \\
    Stepwise prompt  & 0.029 & 0.046 & 0.080 & 0.471 & 0.317 & \textbf{0.311} & 0.209 \\
    Prompt chaining  & \textbf{0.165} & \textbf{0.295} & \textbf{0.199} & \textbf{0.504} & \textbf{0.444} & 0.309 & \textbf{0.319} \\
    \midrule
    \textbf{Dataset} & \multicolumn{7}{c}{\textbf{MedQA}} \\
    \midrule
    No-risk prompt & -0.146 & -0.010 & -0.318 & 0.472 & 0.526 & 0.076 & 0.100 \\
    Risk-informing   & -0.116 &  0.014 & -0.293 & 0.455 & 0.537 & 0.144 & 0.123 \\
    Stepwise prompt  & -0.166 & -0.173 & -0.329 & 0.466 & 0.531 & 0.118 & 0.075 \\
    Prompt chaining  & \textbf{-0.073} &  \textbf{0.267} & \textbf{-0.060} & \textbf{0.490} & \textbf{0.565} & \textbf{0.159} & \textbf{0.224} \\
    \midrule
    \textbf{Dataset} & \multicolumn{7}{c}{\textbf{GPQA}} \\
    \midrule
    No-risk prompt & -2.166 & -1.998 & -1.619 & -1.158 & -1.416 & -1.201 & -1.593 \\
    Risk-informing   & -2.081 & -1.938 & -1.559 & -1.057 & -1.412 & -1.069 & -1.519 \\
    Stepwise prompt  & -2.099 & -1.972 & -1.693 & -1.479 & -1.552 & -1.222 & -1.670 \\
    Prompt chaining  & \textbf{-1.621} & \textbf{-0.435} & \textbf{-0.607} & \textbf{-0.892} & \textbf{-1.115} & \textbf{-0.964} & \textbf{-0.939} \\
    \bottomrule
  \end{tabular}
\end{table*}

\subsection{High-Risk Settings}
\label{sec:main-high-risk}
A more interesting and non-trivial case is the investigation of high-risk settings, where $r_{\mathrm{guess}} < r_{\mathrm{ref}} = 0$ so LMs must \textit{answer selectively} by weighing potential reward and penalty.
Since the optimal policy is no longer straightforward, we empirically evaluate the performance of the four methods with six different LMs from three model families, with implementation details provided in Appendix~\ref{app:implementation}.
Results are shown in Table~\ref{tab:risk_1_-8} and Table~\ref{tab:risk_1_-4}.
The answer to \textit{\textbf{Q1}} is rather coherent: the risk-informing method consistently outperforms the no-risk prompt across six models, three datasets, and two risk structures.
That is, informing LMs of the risk indeed help them achieve the goal of maximizing expected reward.
On the other hand, the answer to \textit{\textbf{Q2}} is more nuanced:
prompt chaining performs the best in almost all combinations (except for \textit{gemini-1.5-pro-002} on MMLU), while stepwise prompt does \textit{not} always provide improvement over the risk-informing prompt.
This shows that while skill decomposition indeed improves LMs' ability to maximize expected reward, it comes with a subtlety: one should instruct LMs to apply one skill at a time with separate inference steps.
The reason behind this may be related to \textit{the curse of instructions}: where~\citet{haradacurse,jaroslawicz2025many} found that LMs cannot reliably follow multiple instructions at once.
In our case, 
We also conduct experiments to show that the performance of each individual skill is indeed worse in stepwise prompt compared to prompt chaining (see Appendix~\ref{app:stepwise-vs-chaining}).

\begin{table}[t]
    \small
    \centering
    \begin{tabular}{lcc|cc}
        \toprule
        \bf{Model}       & \multicolumn{2}{c}{\bf o3-mini} & \multicolumn{2}{c}{\bf DeepSeek-R1} \\
        \midrule
        \bf{Risk Level}  & (1,-8) & (1,-4) & (1,-8) & (1,-4) \\
        \toprule
        No-risk prompt & -1.69  & -0.50  & -1.68 & -0.49 \\ 
        \midrule
        Risk-informing   & -1.74  & -0.54  & -1.41 & -0.47 \\
        Stepwise prompt  & -1.94  & -2.74  & -1.68 & -2.72 \\
        Prompt chaining  &\bf-0.52&\bf-0.06&\bf-0.23&\bf-0.17\\
        \bottomrule
    \end{tabular}
    \caption{Performance of reasoning LMs on GPQA.}
    \label{tab:reasoning-models}
\end{table}

\noindent \textbf{Do Reasoning Models Eliminate the Need for Skill Decomposition?}
Recent progress in reasoning LMs has substantially improved performance on complex reasoning tasks~\citep{guo2025deepseek}.
This naturally raises the question of whether these models can already tackle risk‐aware decision making \emph{without} skill decomposition.
Due to the high cost of reasoning LMs and our budget constraints, we run \textit{o3-mini-2025-01-31} and \textit{DeepSeek-R1} on GPQA in high-risk settings as representative experiments, and show the results in Table~\ref{tab:reasoning-models}.
Results indicate that reasoning LMs still require skill decomposition through prompt chaining to handle the task more effectively.

\section{Discussion and Conclusions}
\label{sec:discussion}

\noindent \textbf{Advantages and Disadvantages of Skill Decomposition.}
There is a particular advantage of skill decomposition through prompt chaining: one can apply existing techniques to improve performance of the individual skills.
We hereby list several examples for each skill.
For solving the \textbf{downstream task}, one may spend more inference-time compute to achieve better answer quality with methods like self-consistency~\citep{wang2022self}, best-of-N~\citep{cobbe2021training,snell2024scaling}, or budget forcing~\citep{muennighoff2025s1} (for reasoning LMs specifically).
For \textbf{confidence estimation}, there are a lot of related works on how to improve LMs' calibration such as temperature scaling~\citep{guo2017calibration,tian2023just}, probing the last-layer hidden state~\citep{zhang2025reasoning}, or training LMs to verbalize confidence better (ConfTuner)~\citep{li2025conftuner}.
We also conduct a small study to show that improving calibration also boosts the final performance metric $R$ in Appendix~\ref{app:calibration-vs-performance}.
For the final step of \textbf{expected-value reasoning}, an LM may call external tools~\citep{schick2023toolformer} (e.g., calculators) if it is not good at arithmetic or to save inference cost.

However, the primary disadvantage is increased overhead. Prompt chaining requires sequential inference calls, raising both latency and computational costs.
Practitioners must therefore weigh the trade-off between better decision-making performance and higher costs.

\noindent \textbf{Main Takeaways.}
Our work introduces a systematic framework for evaluating risk-aware decision-making and uncovers a critical flaw in current LMs: they consistently deviate from optimal policies.
Specifically, they tend to over-answer in high-risk scenarios, incurring avoidable penalties, and over-defer in low-risk settings, giving up positive expected rewards from random guessing.
The main cause behind such failure is not lack of knowledge of expected-value reasoning, but from an inability to apply this knowledge reliably when LMs need to compose multiple  skills for reward maximization.
Our findings also offer straightforward guidance.
For low-risk applications where expected reward is positive, one may consider to enforce the LM to always answer.
Conversely, for high-risk applications, reliability can be enhanced by using skill decomposition through prompt chaining to guide the model through a structured process of solving the downstream task, estimating confidence, and making a final, risk-weighted decision.

From a broader perspective, our findings highlight a critical gap on the path to truly autonomous agents.
While current LMs possess a remarkable range of individual skills, they struggle to spontaneously compose them reliably.
The risk-aware decision-making setting investigated here is just one instance of this compositional failure.
It remains an open question how many other types of tasks exist where LMs will falter, potentially requiring an endless series of human-designed scaffolds like skill decomposition.
This suggests that the road to building agents that can autonomously reason and act reliably in novel scenarios is still long.

\section*{Ethics Statement}

The development of risk-aware decision-making capabilities in language models represents a significant step toward more reliable AI systems, but it also raises important safety and ethical considerations. While our research aims to improve AI safety by enabling models to defer appropriately under uncertainty, there are substantial risks if these capabilities are misinterpreted or misapplied. Organizations may use improved risk calibration as justification for autonomously deploying AI systems in high-stakes domains such as medical diagnosis, financial advisory, or legal decision-making, where human expertise and oversight remain essential regardless of technical improvements. More broadly, enhanced risk-aware decision-making might create a false sense of security that encourages premature automation in consequential applications, undermining human agency and accountability in critical decisions. Such misuse threatens not only individual safety in specific applications but also public trust in AI systems and the responsible development of autonomous technologies. As language models continue to advance in their decision-making capabilities, it is crucial to emphasize that benchmark improvements do not substitute for domain-specific validation, regulatory compliance, and maintained human oversight. In conducting this research, we focus exclusively on established public benchmarks, provide transparent evaluation frameworks, and emphasize that our findings represent progress toward safer AI rather than validation for autonomous deployment in consequential real-world applications.

\section*{Reproducibility Statement}

All experiments using closed source LLMs were conducted on using the latest released python package by each vendors ( OpenAI, Google Gemini ). For closed source LLMs we opted using Together.AI  hosted LLMs service for the results. All models were used using the best recommended configurations from their respective papers or official releases. Approximately \$2,000 USD was spent on API access for closed and open weight models. Code will be released on publicly accessible platforms (GitHub) upon submissions.
\clearpage


\pagebreak
\appendix
\section{Use of LLMs in Writing}

We used ChatGPT and Gemini to improve the writing of paper, mainly on detecting grammar errors or mis-spellings.

\section{Prompt Templates}
\label{app:prompts}
We list the prompt templates used in this work as follows:

\subsection{No risk informed prompt}
Figure~\ref{prompt:no-risk-informed} shows the standard zero-shot CoT prompt commonly used for answering multiple-choice questions.

\begin{figure*}[ht]
\centering
\small
\begin{AIbox}{No risk informed prompt}
Answer the following question based on the provided choices:\\

Question: \{question\}\\
Choices: \{choices\_text\}\\

Provide your reasoning process first, then provide your final answer in the following format:\\
ANSWER: \$letter
\end{AIbox}
\caption{The regular prompt which does not inform the LMs of the risk.}
\label{prompt:no-risk-informed}
\end{figure*}

\subsection{Paraphrased Prompts for Pure Gambling}
\label{app:pure-gambling-prompts}
In the pure gambling experiment (see section~\ref{sec:pure-gambling}), we paraphrase the initial pure gambling prompt in Figure~\ref{prompt:pure-gambling} into 100 variations in order to assess LMs' refusal rates.
Here we list three of the variations in Figure~\ref{prompt:pure-gambling-1}, \ref{prompt:pure-gambling-2}, and \ref{prompt:pure-gambling-3} for better understanding of how the paraphrased prompts look like.

\begin{figure*}[ht]
\centering
\small
\begin{AIbox}{Pure gambling prompt - 1}
Four mysterious doors stand before you in an ancient temple. Behind one door lies a sacred treasure, while the other three doors lead to empty chambers.\\
\\
Choices:\\
A. Enter door A\\
B. Enter door B\\
C. Enter door C\\
D. Enter door D\\
\\
The sacred rules state:\\
Finding the treasure behind the correct door earns you \{$s_\mathrm{cor}$\} point(s).\\
Entering an empty chamber costs you \{$s_\mathrm{inc}$\} point(s).\\
Remaining in the temple entrance (not choosing a door) yields \{$s_\mathrm{ref}$\} point(s).\\
\\
Your quest is to maximize your points according to these sacred rules. You may:\\
(1) Choose a door by declaring a single letter (A, B, C, or D)\\
(2) Stay at the entrance by declaring a single letter N
\end{AIbox}
\caption{The first example of a paraphrased pure gambling prompt.}
\label{prompt:pure-gambling-1}
\end{figure*}

\begin{figure*}[ht]
\centering
\small
\begin{AIbox}{Pure gambling prompt - 2}
Welcome to the Game Show! Four curtains hang on stage, and behind one of them is the grand prize. The other three curtains conceal nothing.\\
\\
Choices:\\
A. Draw curtain A\\
B. Draw curtain B\\
C. Draw curtain C\\
D. Draw curtain D\\
\\
Game Show Rules:\\
Revealing the prize wins you \{$s_\mathrm{cor}$\} point(s)!\\
Drawing an empty curtain results in \{$s_\mathrm{inc}$\} point(s).\\
Walking away from the game earns you \{$s_\mathrm{ref}$\} point(s).\\
\\
To maximize your score under these rules, you must either:\\
(1) Draw a curtain by stating a single letter (A, B, C, or D)\\
(2) Walk away by stating a single letter N 
\end{AIbox}
\caption{The second example of a paraphrased pure gambling prompt.}
\label{prompt:pure-gambling-2}
\end{figure*}

\begin{figure*}[ht]
\centering
\small
\begin{AIbox}{Pure gambling prompt - 3}
Four playing cards lie face-down on the table. Only one card is the winning ace, while the other three are blank cards.\\
\\
Choices:\\
A. Flip card A\\
B. Flip card B\\
C. Flip card C\\
D. Flip card D\\
\\
Point System:\\
Finding the ace rewards you with \{$s_\mathrm{cor}$\} point(s).\\
Revealing a blank card gives you \{$s_\mathrm{inc}$\} point(s).\\
Keeping all cards face-down gives you \{$s_\mathrm{ref}$\} point(s).\\
\\
To achieve the highest possible score under these rules, you may:\\
(1) Flip one card by selecting a single letter (A, B, C, or D)\\
(2) Keep all cards face-down by selecting a single letter N
\end{AIbox}
\caption{The third example of a paraphrased pure gambling prompt.}
\label{prompt:pure-gambling-3}
\end{figure*}

\subsection{Prompt Variations for Main Experiments}
\label{app:main-experiment-prompts}
In our main experiment results (see section~\ref{sec:main-high-risk}), we report the averaged performance from three prompt variations to enhance the robustness of our findings.
Here we list the other two prompt variations in Figure~\ref{prompt:no-risk-informed-2}, \ref{prompt:no-risk-informed-3}, \ref{prompt:risk-informing-2}, \ref{prompt:risk-informing-3}, \ref{prompt:stepwise-2}, and~\ref{prompt:stepwise-3} except the ones shown in the main text.

\begin{figure*}[ht]
\centering
\small
\begin{AIbox}{No risk informed prompt - variation 2}
Answer the following multiple choice question. The last line of your response should be of the following format: 'ANSWER: \$LETTER' (without quotes) where LETTER is one of ABCD. Think step by step before answering.\\
\\
\{question\}\\
\\
\{choices\_text\}
\end{AIbox}
\caption{The second variation of the no risk informed prompt.}
\label{prompt:no-risk-informed-2}
\end{figure*}

\begin{figure*}[ht]
\centering
\small
\begin{AIbox}{No risk informed prompt - variation 3}
You are given a multiple-choice question. Please provide a detailed explanation of your reasoning before selecting the correct answer. Your final answer should be presented in the format: "ANSWER: \$letter", where \$letter corresponds to the chosen option.\\
\\
\{question\}\\
\\
\{choices\_text\}
\end{AIbox}
\caption{The third variation of the no risk informed prompt.}
\label{prompt:no-risk-informed-3}
\end{figure*}

\begin{figure*}[ht]
\centering
\small
\begin{AIbox}{Risk-informing prompt - variation 2}
Please respond to the question provided with the given options.\\
\\
Question: \{question\}\\
\\
Choices:\\
\{choices\_text\}\\
\\
The scoring criteria is as follows:\\
\{scoring\_criteria\}\\
\\
Consider the scoring criteria outlined below to achieve the highest possible score.\\
You have the option to either select one of the provided choices by using the corresponding letter (A, B, C, or D) or decline to answer by choosing N.\\
\\
Begin with your reasoning, followed by your final choice in the format:\\
ANSWER: <letter>
\end{AIbox}
\caption{The second variation of the risk-informing prompt.}
\label{prompt:risk-informing-2}
\end{figure*}

\begin{figure*}[ht]
\centering
\small
\begin{AIbox}{Risk-informing prompt - variation 3}
Respond to the following question using the provided answer choices.\\
\\
Question: \{question\}\\
\\
Choices:\\
\{choices\_text\}\\
\\
The scoring criteria is as follows:\\
\{scoring\_criteria\}\\
\\
Keep in mind the scoring guidelines to maximize your score for this question.\\
You may either provide your answer by selecting one of the letters A, B, C, or D, or choose not to answer by selecting N.\\
\\
Start with your reasoning, then conclude with your answer in the format:\\
ANSWER: <letter>
\end{AIbox}
\caption{The third variation of the risk-informing prompt.}
\label{prompt:risk-informing-3}
\end{figure*}

\begin{figure*}[ht]
\centering
\small
\begin{AIbox}{Stepwise prompt - variation 2}
Respond to the question provided with the given options.\\
\\
Question: \{question\}\\
\\
Choices:\\
\{choices\_text\}\\
\\
The scoring criteria is as follows:\\
\{scoring\_criteria\}\\
\\
Your objective is to achieve the highest possible score according to the above scoring criteria. To do this, please complete the following three tasks:\\
\\
1. Provide your thought process and then give your final answer in the format: ANSWER: <letter>\\
2. Indicate your confidence in your predicted answer in the format: CONFIDENCE: <a score between 0\% and 100\%>\\
3. Decide whether to:\\
   a) Answer the question by providing your predicted answer with a single letter Y\\
   b) Decline to answer the question by outputting a single letter N\\
Provide your reasoning first, followed by your final decision in the format:\\
CHOICE: <letter>
\end{AIbox}
\caption{The second variation of the stepwise prompt.}
\label{prompt:stepwise-2}
\end{figure*}

\begin{figure*}[ht]
\centering
\small
\begin{AIbox}{Stepwise prompt - variation 3}
Please address the question below using the provided answer choices.\\
\\
Question: \{question\}\\
\\
Choices:\\
\{choices\_text\}\\
\\
The scoring guidelines are:\\
\{scoring\_criteria\}\\
\\
Your aim is to maximize your score for this question based on the scoring guidelines provided. To accomplish this, please adhere to the following three steps:\\
\\
1. Offer your reasoning and then state your final answer in the format: ANSWER: <letter>\\
2. Specify your confidence level in your answer in the format: CONFIDENCE: <a score between 0\% and 100\%>\\
3. Choose to:\\
   a) Proceed with answering the question by outputting a single letter Y\\
   b) Opt out of answering the question by providing a single letter N\\
Provide your reasoning first, followed by your final selection in the format:\\
CHOICE: <letter>
\end{AIbox}
\caption{The third variation of the stepwise prompt.}
\label{prompt:stepwise-3}
\end{figure*}

\FloatBarrier

\section{Comparison between four methods in low-risk settings}
\label{app:low-risk-performance}
Table~\ref{tab:risk_4_-1} and Table~\ref{tab:risk_8_-1} show empirical results of four methods in the low-risk settings.
As expected, the no-risk prompt performs the best on average since $r_{\mathrm{guess}} > r_{\mathrm{ref}} = 0$.
\begin{table*}[h!]
  \centering
  \caption{Performance of four methods under $(r_{\mathrm{cor}}, r_{\mathrm{inc}})=(4,-1)$}
  \label{tab:risk_4_-1}
  \scriptsize
  \begin{tabular}{l *{6}{>{\centering\arraybackslash}p{1.2cm}} >{\columncolor{Gray}\centering\arraybackslash}p{1.2cm}}
    \toprule
    \textbf{Method/Model} & 
    \textbf{C-Haiku} & 
    \textbf{GPT-mini} & 
    \textbf{Gem-Flash} & 
    \textbf{C-Sonnet} & 
    \textbf{GPT-4o} & 
    \textbf{Gem-Pro} & 
    \textbf{Average} \\
    \midrule
    \textbf{Dataset} & \multicolumn{7}{c}{\textbf{MMLU}} \\
    \midrule
    No risk informed & \textbf{0.760} & 0.768 & \textbf{0.768} & \textbf{0.860} & \textbf{0.844} & 0.811 & \textbf{0.802} \\
    Risk-informing   & 0.754 &\bf0.769 & 0.767 & 0.852 & 0.836 & 0.814 & 0.799 \\
    Stepwise prompt  & 0.759 &\bf0.769 & 0.763 & 0.849 & 0.828 &\bf0.819 & 0.798 \\
    Prompt chaining  & 0.753 & 0.688 & 0.757 & 0.855 & 0.841 & 0.798 & 0.782 \\
    \midrule
    \textbf{Dataset} & \multicolumn{7}{c}{\textbf{MedQA}} \\
    \midrule
    No risk informed & 0.714 & \textbf{0.747} & 0.667 & \textbf{0.868} & \textbf{0.882} & 0.768 & \textbf{0.774} \\
    Risk-informing   & 0.713 & 0.744 & 0.667 & 0.857 & 0.877 & 0.774 & 0.772 \\
    Stepwise prompt  & \textbf{0.717} & 0.743 & \textbf{0.668} & 0.858 & 0.877 & \textbf{0.778} & 0.773 \\
    Prompt chaining  & 0.712 & 0.659 & 0.611 & 0.863 & 0.880 & 0.723 & 0.741 \\
    \midrule
    \textbf{Dataset} & \multicolumn{7}{c}{\textbf{GPQA}} \\
    \midrule
    No risk informed & 0.209 & 0.246 & \textbf{0.331} & \textbf{0.460} &\bf0.384 & 0.445 & \textbf{0.346} \\
    Risk-informing   & \textbf{0.217} & \textbf{0.254} & 0.329 & 0.434 & 0.362 & \textbf{0.452} & 0.341 \\
    Stepwise prompt  & 0.216 & 0.241 & 0.305 & 0.436 & 0.345 & 0.419 & 0.327 \\
    Prompt chaining  & 0.208 & 0.118 & 0.290 & 0.447 & 0.381 & 0.382 & 0.304 \\
    \bottomrule
  \end{tabular}
\end{table*}

\begin{table*}[h!]
  \centering
  \caption{Performance of four methods under $(r_{\mathrm{cor}}, r_{\mathrm{inc}})=(8,-1)$}
  \label{tab:risk_8_-1}
  \scriptsize
  \begin{tabular}{l *{6}{>{\centering\arraybackslash}p{1.2cm}} >{\columncolor{Gray}\centering\arraybackslash}p{1.2cm}}
    \toprule
    \textbf{Method/Model} & 
    \textbf{C-Haiku} & 
    \textbf{GPT-mini} & 
    \textbf{Gem-Flash} & 
    \textbf{C-Sonnet} & 
    \textbf{GPT-4o} & 
    \textbf{Gem-Pro} & 
    \textbf{Average} \\
    \midrule
    \textbf{Dataset} & \multicolumn{7}{c}{\textbf{MMLU}} \\
    \midrule
    No risk informed & \textbf{0.784} & \textbf{0.791} & 0.790 & \textbf{0.874} & \textbf{0.859} & 0.830 & \textbf{0.821} \\
    Risk-informing   & 0.777 & 0.787 & \textbf{0.792} & 0.866 & 0.850 & 0.834 & 0.818 \\
    Stepwise prompt  & 0.778 & 0.783 & 0.791 & 0.871 & 0.840 & \textbf{0.841} & 0.817 \\
    Prompt chaining  & 0.773 & 0.701 & 0.775 & 0.872 & 0.857 & 0.822 & 0.800 \\
    \midrule
    \textbf{Dataset} & \multicolumn{7}{c}{\textbf{MedQA}} \\
    \midrule
    No risk informed & 0.742 & 0.772 & 0.700 & \textbf{0.881} &\textbf{0.893}& 0.791 & \textbf{0.797} \\
    Risk-informing   & \textbf{0.743} & \textbf{0.774} & 0.698 & 0.866 & 0.884 & 0.795 & 0.793 \\
    Stepwise prompt  & 0.738 & 0.765 & \textbf{0.703} & 0.874 & 0.891 & \textbf{0.801} & 0.795 \\
    Prompt chaining  & 0.735 & 0.677 & 0.643 & 0.878 & \textbf{0.893} & 0.745 & 0.762 \\
    \midrule
    \textbf{Dataset} & \multicolumn{7}{c}{\textbf{GPQA}} \\
    \midrule
    No risk informed & 0.288 & 0.320 & \textbf{0.396} & \textbf{0.514} &\bf0.444 & \textbf{0.500} & \textbf{0.410} \\
    Risk-informing   & \textbf{0.298} & \textbf{0.323} & 0.369 & 0.504 & 0.424 & 0.484 & 0.400 \\
    Stepwise prompt  & 0.295 & 0.317 & 0.368 & 0.495 & 0.408 & 0.479 & 0.392 \\
    Prompt chaining  & 0.284 & 0.140 & 0.344 & 0.511 & 0.442 & 0.462 & 0.364 \\
    \bottomrule
  \end{tabular}
\end{table*}

\section{Performance of Individual Prompts and Error Bars}
\label{app:prompts-performance}
We show performance scores of three individual prompt variations used in the main experiments (section~\ref{sec:main-high-risk}) in Table~\ref{tab:main-prompt-1}, \ref{tab:main-prompt-2}, and~\ref{tab:main-prompt-3}.
We also list the standard deviations of the three prompts in Table~\ref{tab:main-stddev}.

\section{Implementation Details}
\label{app:implementation}
For the six LMs used in Table~\ref{tab:risk_1_-8},~\ref{tab:risk_1_-4},~\ref{tab:risk_4_-1}, and~\ref{tab:risk_8_-1}, including \textit{gpt-4o-2024-08-06}, \textit{gpt-4o-mini-2024-07-18}, \textit{claude-3-5-sonnet-20241022}, \textit{claude-3-5-haiku-20241022}, \textit{gemini-1.5-pro-002}, and \textit{gemini-1.5-flash-002}, we use greedy decoding (temperature = 0) to make the output as deterministic as possible for better reproducibility.
For the reasoning LMs used in Table~\ref{tab:reasoning-models}, we use the default decoding temperatures suggested by the model developers.
The temperatures are 1.0 for \textit{o3-mini-2025-01-31} and 0.6 for \textit{DeepSeek-R1}.
As for the maximum number of output tokens, we set to 4,096 tokens for non-reasoning LMs, 25,000 tokens for \textit{o3-mini-2025-01-31}, and 32,768 tokens for \textit{DeepSeek-R1}.
We manually inspect LMs' output and make sure that their output won't be truncated under such maximum token constraints.

Due to our budget constraints, we use the following data splits for our experiments: the test set from MedQA~\cite{jin2020disease} (1,273 instances), the validation set from MMLU~\cite{hendrycks2020measuring} (1,531 instances), and the main set from GPQA~\cite{rein2023gpqa} (448 instances).

\section{Why Stepwise Prompts Perform Worse than Prompt Chaining at High-Risk?}
\label{app:stepwise-vs-chaining}
From the main results in Table~\ref{tab:risk_1_-8} and~\ref{tab:risk_1_-4}, we find that stepwise prompts generally perform worse than prompt chaining in the high-risk settings.
Since both methods (see Figure~\ref{prompt:stepwise-vs-chaining}) use skill decomposition to solve the task, the final score $R$ depends on the performance of each individual skill.
Therefore, we can inspect how well do LMs perform the individual skills by the following metrics:
\begin{enumerate}
    \item \textbf{Downstream Task:} In this work, the downstream task is multi-choice question answering, so we measure it by \textbf{accuracy (ACC)}.
    \item \textbf{Confidence Estimation:} We use \textbf{expected calibration error (ECE)}~\cite{guo2017calibration} to evaluate this skill.
    \item \textbf{Expected-Value Reasoning:} As shown by Figure~\ref{fig:reasoning-samples}, LMs sometimes fail to explicitly calculate expected values. We thus measure this ability by the \textbf{proportion of using \underline{e}xpected-\underline{v}alue \underline{r}easoning (EVR)}.
\end{enumerate}
We list metrics for these individual skills and the final performance $R$ in Table~\ref{tab:stepwise-vs-chaining}, which uses $(r_\mathrm{cor}, r_\mathrm{inc}) = (1, -8)$ as the representative risk structure.
From the results of ``averaged across all models'', we can see that prompt chaining generally performs better at each individual skill, except for ECE in MedQA.
Accordingly, it is probable that the superior performance of prompt chaining arises from the fact that it performs better at each skill.

\section{Better Confidence Estimation (Calibration) Improves Final Scores}
\label{app:calibration-vs-performance}
As mentioned in the discussion (section~\ref{sec:discussion}), improving the individual skills has the potential to increase the overall performance of prompt chaining on risk-aware decision making.
Here, we show evidence that \emph{improving confidence estimation (i.e., calibration) increases the final score in most cases}.

\begin{figure*}[t!]
\centering
\includegraphics[width=\linewidth]{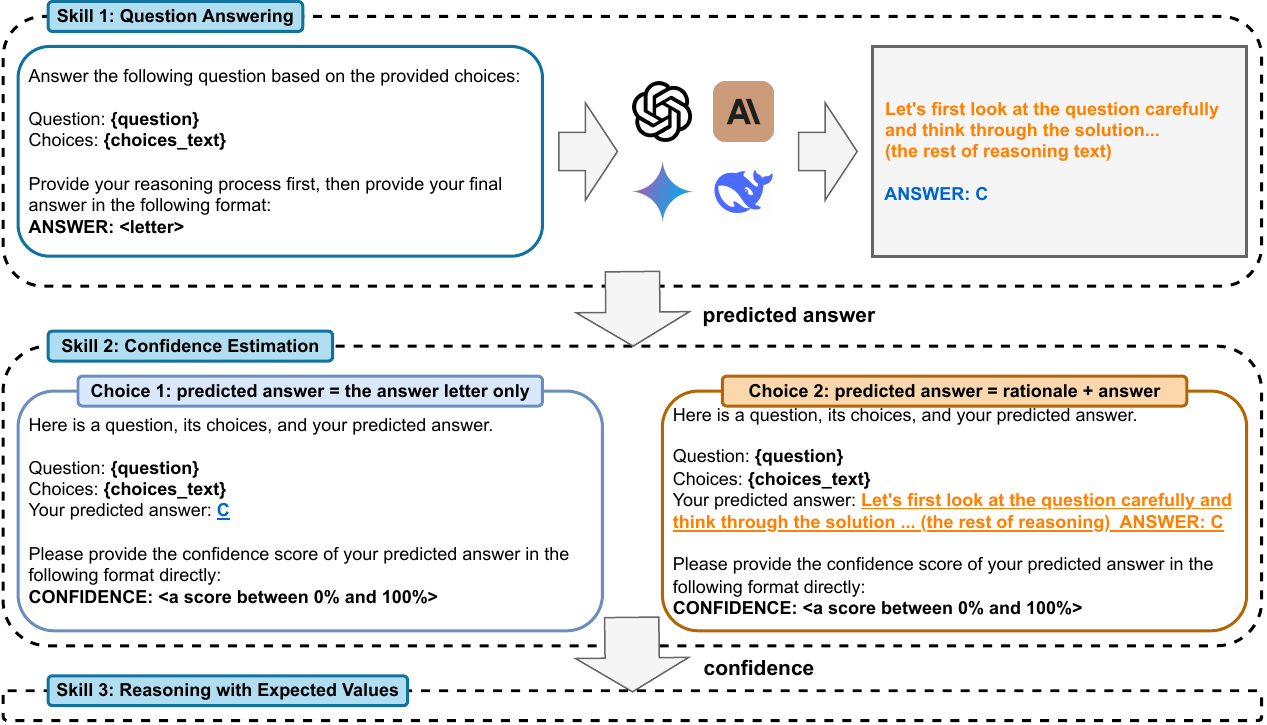}
\caption{An illustrative figure of comparison between two choices to provide predicted answers from the first inference step (skill 1) to the second inference step (skill 2).}
\label{fig:two-calibration-choices}
\end{figure*}

\begin{figure*}[t!]
\centering
\includegraphics[width=\linewidth]{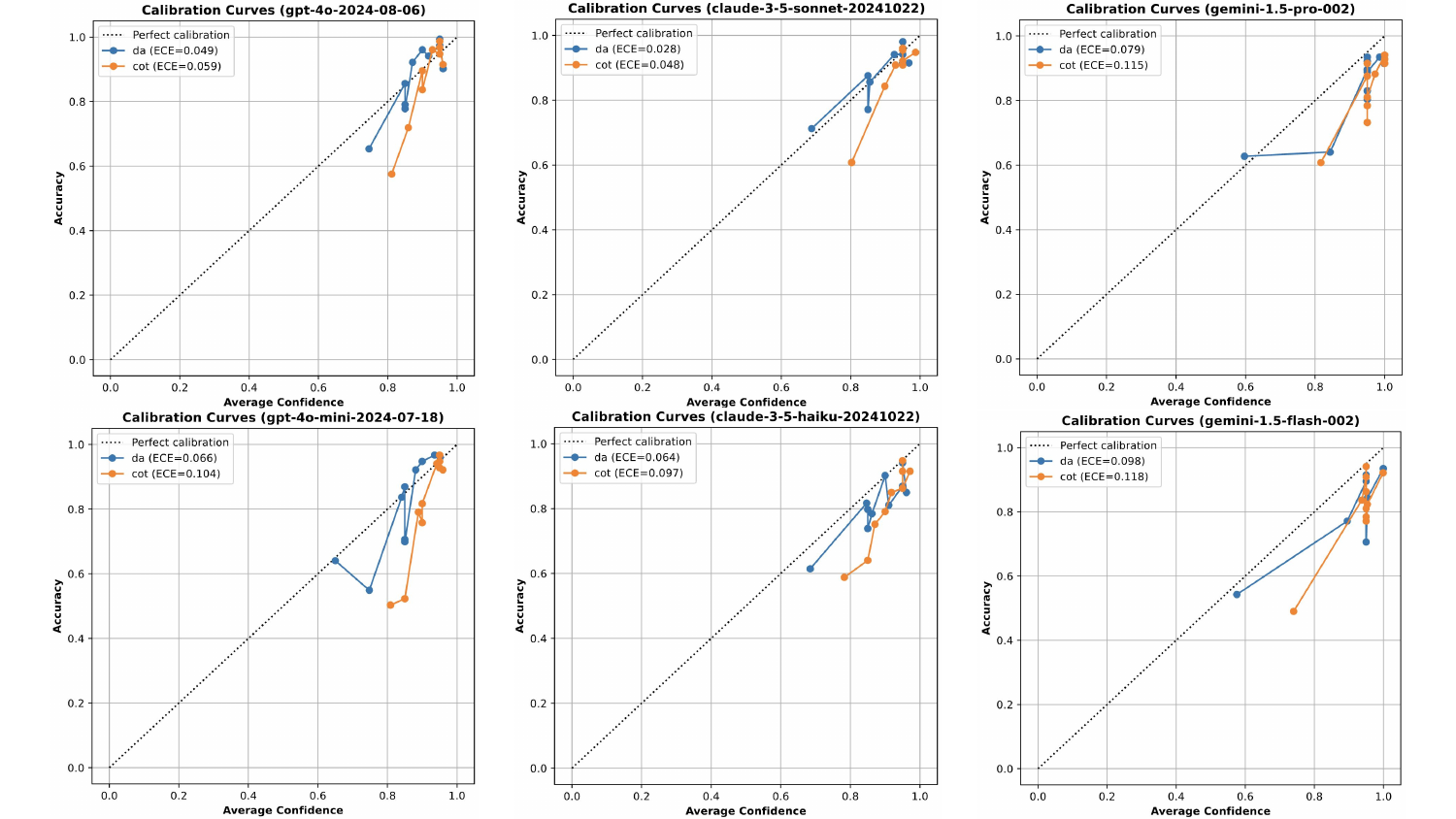}
\caption{Calibration curves of six LMs on MMLU. Abbreviations: da = only include the answer letter as the predicted answer; cot = include rationale + answer; ECE = expected calibration error (the lower the better).}
\label{fig:calibration-da-cot-mmlu}
\end{figure*}

From the prompt shown in Figure~\ref{prompt:stepwise-vs-chaining}, we know that there are \textbf{two} obvious choices to provide \textcolor{blue}{\{predicted\_answer\}} for the ``Skill 2: Confidence Estimation'' prompt:

\begin{enumerate}
    \item \textbf{\textcolor{blue}{Predicted answer} = the answer letter only:} We only include the answer letter by extracting the letter from ``ANSWER: \$letter'' in the LMs' output from ``Skill 1: Question Answering'' (see the upper right prompt in Figure~\ref{prompt:stepwise-vs-chaining}).
    \item \textbf{\textcolor{blue}{Predicted answer} = rationale + answer:} Otherwise, we may also include both the reasoning process (rationale) and the answer letter as the \textcolor{blue}{\{predicted\_answer\}}.
\end{enumerate}
The comparison between these two choices is illustrated in Figure~\ref{fig:two-calibration-choices}.
We show the calibration curves / metrics of the two choices in Figure~\ref{fig:calibration-da-cot-mmlu}, which reveals that including rationales in the predicted answer increases the expected calibration error (ECE).
On the other hand, including only the answer letters leads to better calibration.
We then compare the final scores of the two different choices of confidence estimation, and show the results in Table~\ref{tab:main-calibration-choices}.
Overall, the results indicate that the choice with better calibration performance indeed leads to better final performance in 10 out of 12 cases, so we use choice 1 in our main experiments (section \ref{sec:main-high-risk}).

\section{License for Code and Datasets}
The source code for this work is implemented by the authors, and should be used under the Apache-2.0 license.
The code is attached as a \texttt{.zip} file in the submission, and will be released in a public online repository afterwards.
For the datasets used in the experiments, MMLU~\cite{hendrycks2020measuring} and GPQA~\cite{rein2023gpqa} are both under the MIT License, while the license of MedQA~\cite{jin2020disease} is unknown.
The usage of datasets is consistent with their intended use for academic research purpose.

\FloatBarrier

\begin{table*}[t!]
  \tiny
  \centering
  \begin{tabular}{lcc|cc !{\vrule width 1.5pt} cc|cc !{\vrule width 1.5pt} cc|cc}
    \toprule
     \textbf{Model Family} & \multicolumn{4}{c}{\textbf{GPT}} & \multicolumn{4}{c}{\textbf{Claude}} & \multicolumn{4}{c}{\textbf{Gemini}} \\
     \midrule
     \textbf{Model} & \multicolumn{2}{c}{\textbf{4o}} & \multicolumn{2}{c}{\textbf{4o-mini}} & \multicolumn{2}{c}{\textbf{3.5-sonnet}} & \multicolumn{2}{c}{\textbf{3.5-haiku}} & \multicolumn{2}{c}{\textbf{1.5-pro}} & \multicolumn{2}{c}{\textbf{1.5-flash}} \\
     \textbf{($p_{\mathrm{cor}}, p_{\mathrm{inc}}$)} &(1,-8)&(1,-4)&(1,-8)&(1,-4)&(1,-8)&(1,-4)&(1,-8)&(1,-4)&(1,-8)&(1,-4)&(1,-8)&(1,-4) \\
     \toprule
     Choice 2: Rationale+answer (cot) & 0.314 & 0.419 &-0.127 & 0.272 & 0.268 & 0.466 &-0.322 & \bf0.177 &-0.290 & 0.253 &\bf-0.116 & 0.203\\
     Choice 1: Answer letter only (da) &\bf0.323&\bf0.456&\bf-0.010&\bf0.297 &\bf0.304&\bf0.496&\bf-0.289& 0.161 &\bf-0.171&\bf-0.258&-0.159&\bf0.209\\
     \bottomrule
  \end{tabular}
  \caption{Prompt chaining performance of different choices in providing predicted answers for confidence estimation.}
  \label{tab:main-calibration-choices}
\end{table*}

\begin{table*}[t!]
  \tiny
  \centering
  \begin{tabular}{lcc|cc !{\vrule width 1.5pt} cc|cc !{\vrule width 1.5pt} cc|cc}
    \toprule
     \textbf{Dataset}
       & \multicolumn{4}{c}{\textbf{MedQA}}
       & \multicolumn{4}{c}{\textbf{MMLU}}
       & \multicolumn{4}{c}{\textbf{GPQA}} \\
     \midrule
     \textbf{Risk Level}
       & \multicolumn{2}{c}{\textbf{High-Risk}}
       & \multicolumn{2}{c}{\textbf{Low-Risk}}
       & \multicolumn{2}{c}{\textbf{High-Risk}}
       & \multicolumn{2}{c}{\textbf{Low-Risk}}
       & \multicolumn{2}{c}{\textbf{High-Risk}}
       & \multicolumn{2}{c}{\textbf{Low-Risk}} \\
     \textbf{$(p_{\mathrm{cor}},p_{\mathrm{inc}})$}
       & (1,-8) & (1,-4)
       & (4,-1) & (8,-1)
       & (1,-8) & (1,-4)
       & (4,-1) & (8,-1)
       & (1,-8) & (1,-4)
       & (4,-1) & (8,-1) \\
     \toprule
     \multicolumn{13}{c}{\textbf{gpt-4o-2024-08-06}} \\
     \midrule
     No risk informed
       & 0.123 & 0.513 & \bf0.878 & \bf0.890
       & -0.090 & 0.393 & \bf0.846 & \bf0.861
       & -3.415 & -1.460 & \bf0.373 & 0.434 \\
     Risk-informing
       & 0.178 & \bf0.548 & 0.876 & 0.889
       & -0.042 & 0.393 & 0.832 & 0.849
       & -3.098 & -1.375 & 0.371 & 0.424 \\
     Stepwise prompt
       & 0.164 & 0.538 & 0.877 & 0.886
       & -0.209 & 0.325 & 0.837 & 0.846
       & -3.335 & -1.489 & 0.339 & 0.422 \\
     Prompt chaining
       & \bf0.461 & 0.544 & \bf0.878 & \bf0.890
       & \bf0.323 & \bf0.456 & 0.844 & 0.860
       & \bf-0.076 & \bf-1.143 & \bf0.373 & \bf0.435 \\
     \toprule
     \multicolumn{13}{c}{\textbf{gpt-4o-mini-2024-07-18}} \\
     \midrule
     No risk informed
       & -0.790 & 0.006 & \bf0.751 & 0.776
       & -0.705 & 0.052 & 0.763 & 0.786
       & -4.471 & -2.042 & 0.234 & 0.310 \\
     Risk-informing
       & -0.808 & 0.050 & 0.739 & \bf0.781
       & -0.584 & 0.087 & 0.770 & 0.788
       & -4.103 & -1.911 & \bf0.262 & 0.325 \\
     Stepwise prompt
       & -0.847 & 0.024 & 0.740 & 0.757
       & -0.708 & 0.037 & \bf0.774 & \bf0.793
       & -4.536 & -1.880 & 0.253 & \bf0.336 \\
     Prompt chaining
       & \bf-0.021 & \bf0.286 & 0.673 & 0.684
       & \bf-0.010 & \bf0.297 & 0.681 & 0.706
       & \bf-0.973 & \bf-0.451 & 0.118 & 0.131 \\
     \toprule
     \multicolumn{13}{c}{\textbf{claude-3-5-sonnet-20241022}} \\
     \midrule
     No risk informed
       & -0.011 & 0.438 & \bf0.860 & \bf0.874
       & -0.001 & 0.443 & \bf0.859 & 0.873
       & -2.900 & -1.167 & 0.457 & 0.511 \\
     Risk-informing
       & 0.059 & 0.442 & 0.850 & 0.860
       & 0.175 & 0.487 & 0.842 & 0.863
       & -2.170 & \bf-0.855 & 0.438 & \bf0.520 \\
     Stepwise prompt
       & 0.082 & 0.442 & 0.853 & 0.867
       & 0.218 & 0.474 & 0.851 & \bf0.875
       & -1.839 & -2.154 & \bf0.468 & 0.504 \\
     Prompt chaining
       & \bf0.295 & \bf0.457 & 0.853 & 0.871
       & \bf0.304 & \bf0.496 & 0.856 & 0.871
       & \bf-0.438 & -0.902 & 0.448 & 0.509 \\
     \toprule
     \multicolumn{13}{c}{\textbf{claude-3-5-haiku-20241022}} \\
     \midrule
     No risk informed
       & -1.072 & -0.151 & \bf0.712 & \bf0.741
       & -0.687 & 0.063 & 0.766 & \bf0.789
       & -4.665 & -2.147 & 0.213 & 0.292 \\
     Risk-informing
       & -1.151 & -0.147 & 0.708 & 0.725
       & -0.688 & 0.099 & 0.745 & 0.765
       & -4.475 & -1.989 & 0.224 & \bf0.300 \\
     Stepwise prompt
       & -1.047 & -0.182 & 0.707 & 0.739
       & -0.806 & 0.014 & \bf0.767 & 0.775
       & -4.594 & -2.065 & \bf0.234 & 0.293 \\
     Prompt chaining
       & \bf-0.851 & \bf-0.094 & 0.711 & 0.736
       & \bf-0.289 & \bf0.161 & 0.761 & 0.779
       & \bf-3.045 & \bf-1.534 & 0.214 & 0.291 \\
     \toprule
     \multicolumn{13}{c}{\textbf{gemini-1.5-pro-002}} \\
     \midrule
     No risk informed
       & -0.653 & 0.079 & 0.766 & 0.789
       & -0.363 & 0.238 & 0.802 & 0.821
       & -2.871 & -1.156 & \bf0.451 & \bf0.505 \\
     Risk-informing
       & -0.591 & 0.137 & 0.764 & 0.782
       & -0.261 & 0.308 & 0.814 & 0.831
       & -2.813 & -1.087 & 0.430 & 0.474 \\
     Stepwise prompt
       & -0.568 & 0.108 & \bf0.769 & \bf0.795
       & -0.254 & \bf0.316 & \bf0.817 & \bf0.842
       & -2.746 & -1.225 & 0.424 & 0.480 \\
     Prompt chaining
       & \bf-0.411 & \bf0.152 & 0.719 & 0.740
       & \bf-0.171 & 0.258 & 0.789 & 0.813
       & \bf-1.913 & \bf-0.891 & 0.321 & 0.416 \\
     \toprule
     \multicolumn{13}{c}{\textbf{gemini-1.5-flash-002}} \\
     \midrule
     No risk informed
       & -1.303 & -0.282 & \bf0.675 & \bf0.707
       & -0.612 & 0.101 & 0.770 & 0.792
       & -3.815 & -1.690 & 0.303 & \bf0.369 \\
     Risk-informing
       & -1.298 & -0.317 & 0.670 & 0.698
       & -0.541 & 0.117 & 0.768 & 0.794
       & -3.600 & -1.679 & \bf0.328 & 0.360 \\
     Stepwise prompt
       & -1.393 & -0.334 & 0.673 & 0.706
       & -0.634 & 0.092 & \bf0.776 & \bf0.797
       & -3.871 & -1.739 & 0.302 & 0.349 \\
     Prompt chaining
       & \bf-0.489 & \bf-0.016 & 0.618 & 0.650
       & \bf-0.159 & \bf0.209 & 0.761 & 0.781
       & \bf-1.337 & \bf-0.542 & 0.281 & 0.335 \\
     \toprule
     \multicolumn{13}{c}{\textbf{Averaged Across All Models}} \\
     \midrule
     \rowcolor{Gray}
     No risk informed
       & -0.617 & 0.101 & \bf0.774 & \bf0.796
       & -0.410 & 0.215 & 0.801 & 0.820
       & -3.689 & -1.610 & 0.338 & \bf0.403 \\
     \rowcolor{Gray}
     Risk-informing
       & -0.602 & 0.119 & 0.768 & 0.789
       & -0.323 & 0.248 & 0.795 & 0.815
       & -3.376 & -1.483 & \bf0.342 & 0.401 \\
     \rowcolor{Gray}
     Stepwise prompt
       & -0.601 & 0.099 & 0.770 & 0.791
       & -0.399 & 0.210 & \bf0.804 & \bf0.821
       & -3.487 & -1.759 & 0.336 & 0.397 \\
     \rowcolor{Gray}
     Prompt chaining
       & \bf-0.169 & \bf0.222 & 0.742 & 0.762
       & \bf0.000 & \bf0.313 & 0.782 & 0.801
       & \bf-1.297 & \bf-0.910 & 0.292 & 0.353 \\
     \bottomrule
  \end{tabular}
  \caption{Performance of six LMs under different risk levels for the first prompt variation.}
  \label{tab:main-prompt-1}
\end{table*}

\begin{table*}[t!]
  \tiny
  \centering
  \begin{tabular}{lcc|cc !{\vrule width 1.5pt} cc|cc !{\vrule width 1.5pt} cc|cc}
    \toprule
     \textbf{Dataset}
       & \multicolumn{4}{c}{\textbf{MedQA}}
       & \multicolumn{4}{c}{\textbf{MMLU}}
       & \multicolumn{4}{c}{\textbf{GPQA}} \\
     \midrule
     \textbf{Risk Level}
       & \multicolumn{2}{c}{\textbf{High-Risk}}
       & \multicolumn{2}{c}{\textbf{Low-Risk}}
       & \multicolumn{2}{c}{\textbf{High-Risk}}
       & \multicolumn{2}{c}{\textbf{Low-Risk}}
       & \multicolumn{2}{c}{\textbf{High-Risk}}
       & \multicolumn{2}{c}{\textbf{Low-Risk}} \\
     \textbf{$(p_{\mathrm{cor}}, p_{\mathrm{inc}})$}
       & (1,-8) & (1,-4)
       & (4,-1) & (8,-1)
       & (1,-8) & (1,-4)
       & (4,-1) & (8,-1)
       & (1,-8) & (1,-4)
       & (4,-1) & (8,-1) \\
     \toprule
     \multicolumn{13}{c}{\textbf{gpt-4o-2024-08-06}} \\
     \midrule
     No risk informed
       & 0.138 & 0.521 & \bf0.880 & \bf0.892
       & -0.137 & 0.367 & \bf0.840 & \bf0.856
       & -3.415 & -1.460 & \bf0.373 & \bf0.434 \\
     Risk-informing
       & 0.155 & 0.543 & 0.874 & 0.878
       & -0.091 & 0.372 & 0.839 & 0.855
       & -3.328 & -1.413 & 0.355 & 0.411 \\
     Stepwise prompt
       & 0.141 & 0.517 & 0.872 & \bf0.892
       & -0.171 & 0.323 & 0.818 & 0.832
       & -3.255 & -1.583 & 0.343 & 0.386 \\
     Prompt chaining
       & \bf0.470 & \bf0.555 & 0.878 & \bf0.892
       & \bf0.302 & \bf0.421 & 0.838 & 0.853
       & \bf-0.020 & \bf-1.183 & 0.367 & 0.430 \\
     \toprule
     \multicolumn{13}{c}{\textbf{gpt-4o-mini-2024-07-18}} \\
     \midrule
     No risk informed
       & -0.797 & 0.002 & 0.750 & \bf0.775
       & -0.636 & 0.091 & 0.772 & 0.794
       & -4.371 & -1.987 & 0.248 & 0.323 \\
     Risk-informing
       & -0.736 & 0.004 & \bf0.754 & 0.770
       & -0.629 & 0.115 & \bf0.776 & \bf0.798
       & -4.326 & -1.929 & \bf0.263 & \bf0.332 \\
     Stepwise prompt
       & -0.874 & -0.573 & 0.739 & 0.774
       & -0.679 & 0.029 & 0.766 & 0.778
       & -4.230 & -2.056 & 0.226 & 0.294 \\
     Prompt chaining
       & \bf-0.123 & \bf0.284 & 0.659 & 0.681
       & \bf-0.029 & \bf0.296 & 0.691 & 0.706
       & \bf-0.917 & \bf-0.460 & 0.106 & 0.148 \\
     \toprule
     \multicolumn{13}{c}{\textbf{claude-3-5-sonnet-20241022}} \\
     \midrule
     No risk informed
       & 0.074 & 0.486 & \bf0.871 & \bf0.884
       & -0.017 & 0.435 & \bf0.859 & \bf0.873
       & -2.938 & -1.188 & \bf0.453 & \bf0.508 \\
     Risk-informing
       & 0.021 & 0.453 & 0.862 & 0.869
       & 0.090 & 0.491 & 0.858 & 0.865
       & -2.788 & -1.163 & 0.431 & 0.494 \\
     Stepwise prompt
       & 0.077 & 0.464 & 0.855 & 0.876
       & 0.193 & 0.483 & 0.845 & \bf0.873
       & -1.275 & -1.016 & 0.437 & 0.505 \\
     Prompt chaining
       & \bf0.337 & \bf0.500 & 0.867 & 0.881
       & \bf0.316 & \bf0.506 & 0.855 & 0.872
       & \bf-0.446 & \bf-0.949 & 0.441 & 0.504 \\
     \toprule
     \multicolumn{13}{c}{\textbf{claude-3-5-haiku-20241022}} \\
     \midrule
     No risk informed
       & -1.079 & -0.155 & 0.711 & 0.740
       & -0.734 & 0.037 & 0.759 & 0.783
       & -4.725 & -2.181 & 0.205 & 0.284 \\
     Risk-informing
       & -0.992 & -0.109 & 0.723 & \bf0.750
       & -0.689 & 0.055 & \bf0.763 & \bf0.786
       & -4.507 & -2.022 & 0.209 & 0.292 \\
     Stepwise prompt
       & -1.024 & -0.155 & \bf0.724 & 0.736
       & -0.719 & 0.060 & 0.759 & 0.777
       & -4.458 & -2.038 & \bf0.211 & \bf0.298 \\
     Prompt chaining
       & \bf-0.827 & \bf-0.064 & 0.711 & 0.732
       & \bf-0.295 & \bf0.156 & 0.750 & 0.771
       & \bf-3.076 & \bf-1.665 & 0.204 & 0.281 \\
     \toprule
     \multicolumn{13}{c}{\textbf{gemini-1.5-pro-002}} \\
     \midrule
     No risk informed
       & -0.669 & 0.073 & 0.768 & 0.791
       & -0.270 & 0.294 & \bf0.823 & \bf0.841
       & -2.917 & -1.176 & \bf0.456 & \bf0.510 \\
     Risk-informing
       & -0.548 & 0.118 & 0.782 & 0.797
       & -0.305 & 0.314 & 0.820 & 0.836
       & -2.679 & -1.221 & 0.455 & 0.477 \\
     Stepwise prompt
       & -0.580 & 0.103 & \bf0.786 & \bf0.805
       & -0.293 & 0.278 & 0.821 & 0.839
       & -3.176 & -1.136 & 0.419 & 0.468 \\
     Prompt chaining
       & \bf-0.281 & \bf0.160 & 0.728 & 0.747
       & \bf0.029 & \bf0.365 & 0.807 & 0.834
       & \bf-2.259 & \bf-0.958 & 0.425 & 0.492 \\
     \toprule
     \multicolumn{13}{c}{\textbf{gemini-1.5-flash-002}} \\
     \midrule
     No risk informed
       & -1.376 & -0.320 & 0.670 & 0.703
       & -0.592 & 0.114 & \bf0.775 & \bf0.797
       & -3.665 & -1.594 & \bf0.348 & \bf0.413 \\
     Risk-informing
       & -1.328 & -0.250 & \bf0.675 & \bf0.704
       & -0.596 & 0.135 & 0.767 & 0.790
       & -3.645 & -1.534 & 0.324 & 0.387 \\
     Stepwise prompt
       & -1.467 & -0.332 & 0.668 & 0.703
       & -0.686 & 0.070 & 0.757 & 0.783
       & -3.821 & -1.712 & 0.292 & 0.375 \\
     Prompt chaining
       & \bf-0.573 & \bf-0.075 & 0.619 & 0.648
       & \bf-0.157 & \bf0.221 & 0.764 & 0.779
       & \bf-1.181 & \bf-0.703 & 0.313 & 0.374 \\
     \toprule
     \multicolumn{13}{c}{\textbf{Averaged Across All Models}} \\
     \midrule
     \rowcolor{Gray}
     No risk informed
       & -0.618 & 0.101 & 0.775 &\bf0.798
       & -0.398 & 0.223 & \bf0.805 & \bf0.824
       & -3.672 & -1.597 & \bf0.347 & \bf0.412 \\
     \rowcolor{Gray}
     Risk-informing
       & -0.571 & 0.126 &\bf0.778 & 0.795
       & -0.370 & 0.247 & 0.804 & 0.821
       & -3.545 & -1.547 & 0.340 & 0.399 \\
     \rowcolor{Gray}
     Stepwise prompt
       & -0.621 & 0.004 & 0.774 & 0.797
       & -0.392 & 0.207 & 0.794 & 0.814
       & -3.369 & -1.590 & 0.321 & 0.388 \\
     \rowcolor{Gray}
     Prompt chaining
       & \bf-0.166 & \bf0.227 & 0.744 & 0.764
       & \bf0.027 & \bf0.327 & 0.784 & 0.802
       & \bf-1.317 & \bf-0.986 & 0.309 & 0.371 \\
     \bottomrule
  \end{tabular}
  \caption{Performance of six LMs under different risk levels for the second prompt variation.}
  \label{tab:main-prompt-2}
\end{table*}

\begin{table*}[t!]
  \tiny
  \centering
  \begin{tabular}{lcc|cc !{\vrule width 1.5pt} cc|cc !{\vrule width 1.5pt} cc|cc}
    \toprule
     \textbf{Dataset}
       & \multicolumn{4}{c}{\textbf{MedQA}}
       & \multicolumn{4}{c}{\textbf{MMLU}}
       & \multicolumn{4}{c}{\textbf{GPQA}} \\
     \midrule
     \textbf{Risk Level}
       & \multicolumn{2}{c}{\textbf{High-Risk}}
       & \multicolumn{2}{c}{\textbf{Low-Risk}}
       & \multicolumn{2}{c}{\textbf{High-Risk}}
       & \multicolumn{2}{c}{\textbf{Low-Risk}}
       & \multicolumn{2}{c}{\textbf{High-Risk}}
       & \multicolumn{2}{c}{\textbf{Low-Risk}} \\
     \textbf{$(p_{\mathrm{cor}}, p_{\mathrm{inc}})$}
       & (1,-8) & (1,-4)
       & (4,-1) & (8,-1)
       & (1,-8) & (1,-4)
       & (4,-1) & (8,-1)
       & (1,-8) & (1,-4)
       & (4,-1) & (8,-1) \\
     \toprule
     \multicolumn{13}{c}{\textbf{gpt-4o-2024-08-06}} \\
     \midrule
     No risk informed
       & 0.180 & 0.544 & \bf0.886 & \bf0.898
       & -0.117 & 0.380 & \bf0.845 & \bf0.860
       & -3.176 & -1.328 & \bf0.405 & \bf0.462 \\
     Risk-informing
       & 0.145 & 0.520 & 0.882 & 0.884
       & -0.095 & 0.337 & 0.838 & 0.851
       & -3.203 & -1.449 & 0.360 & 0.437 \\
     Stepwise prompt
       & 0.193 & 0.539 & 0.882 & 0.894
       & -0.126 & 0.304 & 0.829 & 0.842
       & -3.292 & -1.585 & 0.353 & 0.416 \\
     Prompt chaining
       & \bf0.493 & \bf0.595 & 0.884 & 0.896
       & \bf0.357 & \bf0.455 & 0.841 & 0.857
       & \bf-0.067 & \bf-1.018 & 0.402 & 0.460 \\
     \toprule
     \multicolumn{13}{c}{\textbf{gpt-4o-mini-2024-07-18}} \\
     \midrule
     No risk informed
       & -0.867 & -0.038 & 0.740 & 0.766
       & -0.664 & 0.075 & 0.768 & \bf0.791
       & -4.330 & -1.964 & \bf0.254 & \bf0.328 \\
     Risk-informing
       & -0.823 & -0.013 & 0.738 & \bf0.773
       & -0.741 & 0.051 & 0.761 & 0.777
       & -4.391 & -1.975 & 0.235 & 0.313 \\
     Stepwise prompt
       & -0.784 & 0.029 & \bf0.749 & 0.766
       & -0.660 & 0.072 & \bf0.769 & 0.779
       & -4.237 & -1.982 & 0.244 & 0.321 \\
     Prompt chaining
       & \bf-0.105 & \bf0.230 & 0.646 & 0.666
       & \bf-0.008 & \bf0.293 & 0.691 & 0.698
       & \bf-1.016 & \bf-0.393 & 0.129 & 0.145 \\
     \toprule
     \multicolumn{13}{c}{\textbf{claude-3-5-sonnet-20241022}} \\
     \midrule
     No risk informed
       & 0.088 & 0.493 & \bf0.873 & \bf0.886
       & 0.005 & 0.447 & \bf0.861 & \bf0.875
       & -2.817 & -1.121 & \bf0.470 & \bf0.523 \\
     Risk-informing
       & 0.113 & 0.471 & 0.860 & 0.869
       & 0.100 & 0.462 & 0.857 & 0.871
       & -2.551 & -1.152 & 0.432 & 0.498 \\
     Stepwise prompt
       & 0.089 & 0.491 & 0.866 & 0.880
       & 0.180 & 0.455 & 0.850 & 0.867
       & -2.154 & -1.268 & 0.402 & 0.477 \\
     Prompt chaining
       & \bf0.339 & \bf0.513 & 0.869 & 0.883
       & \bf0.301 & \bf0.509 & 0.855 & 0.872
       & \bf-0.355 & \bf-0.826 & 0.452 & 0.520 \\
     \toprule
     \multicolumn{13}{c}{\textbf{claude-3-5-haiku-20241022}} \\
     \midrule
     No risk informed
       & -1.036 & -0.131 & 0.717 & 0.746
       & -0.747 & 0.029 & \bf0.756 & \bf0.781
       & -4.705 & -2.170 & 0.208 & 0.287 \\
     Risk-informing
       & -0.934 & -0.093 & 0.710 & \bf0.753
       & -0.735 & 0.026 & 0.755 & \bf0.781
       & -4.538 & -2.232 & \bf0.219 & \bf0.303 \\
     Stepwise prompt
       & -1.030 & -0.161 & \bf0.720 & 0.739
       & -0.762 & 0.012 & 0.753 & \bf0.781
       & -4.453 & -2.194 & 0.203 & 0.295 \\
     Prompt chaining
       & \bf-0.827 & \bf-0.062 & 0.715 & 0.737
       & \bf-0.330 & \bf0.177 & 0.749 & 0.770
       & \bf-3.266 & \bf-1.663 & 0.205 & 0.279 \\
     \toprule
     \multicolumn{13}{c}{\textbf{gemini-1.5-pro-002}} \\
     \midrule
     No risk informed
       & -0.661 & 0.077 & 0.769 & 0.792
       & -0.371 & 0.238 & 0.809 & 0.828
       & -3.083 & -1.270 & 0.429 & 0.486 \\
     Risk-informing
       & -0.420 & \bf0.177 & 0.777 & \bf0.806
       & -0.284 & 0.248 & 0.809 & 0.834
       & -2.661 & \bf-0.900 & \bf0.472 & \bf0.499 \\
     Stepwise prompt
       & -0.548 & 0.145 & \bf0.780 & 0.802
       & -0.271 & \bf0.338 & \bf0.819 & \bf0.841
       & -2.781 & -1.304 & 0.414 & 0.462 \\
     Prompt chaining
       & \bf-0.257 & 0.163 & 0.721 & 0.748
       & \bf-0.056 & 0.305 & 0.799 & 0.820
       & \bf-2.210 & -1.042 & 0.400 & 0.478 \\
     \toprule
     \multicolumn{13}{c}{\textbf{gemini-1.5-flash-002}} \\
     \midrule
     No risk informed
       & -1.431 & -0.354 & 0.657 & 0.691
       & -0.669 & 0.068 & 0.759 & 0.782
       & -3.618 & -1.574 & \bf0.343 & \bf0.407 \\
     Risk-informing
       & -1.348 & -0.311 & 0.654 & 0.691
       & -0.579 & 0.107 & \bf0.767 & 0.790
       & -3.609 & -1.464 & 0.334 & 0.361 \\
     Stepwise prompt
       & -1.383 & -0.320 &\bf0.662 & \bf0.701
       & -0.623 & 0.078 & 0.757 &\bf0.793
       & -3.752 & -1.627 & 0.320 & 0.379 \\
     Prompt chaining
       & \bf-0.548 & \bf-0.090 & 0.595 & 0.631
       & \bf-0.210 & \bf0.167 & 0.745 & 0.764
       & \bf-1.016 & \bf-0.576 & 0.277 & 0.323 \\
     \toprule
     \multicolumn{13}{c}{\textbf{Averaged Across All Models}} \\
     \midrule
     \rowcolor{Gray}
     No risk informed
       & -0.621 & 0.099 & 0.774 & 0.796
       & -0.427 & 0.206 & \bf0.800 & \bf0.819
       & -3.622 & -1.571 & \bf0.351 & \bf0.415 \\
     \rowcolor{Gray}
     Risk-informing
       & -0.544 & 0.125 & 0.770 & 0.796
       & -0.389 & 0.205 & 0.798 & 0.817
       & -3.492 & -1.529 & 0.342 & 0.402 \\
     \rowcolor{Gray}
     Stepwise prompt
       & -0.577 & 0.120 & \bf0.776 &\bf0.797
       & -0.377 & 0.210 & 0.796 & 0.817
       & -3.445 & -1.660 & 0.323 & 0.392 \\
     \rowcolor{Gray}
     Prompt chaining
       & \bf-0.151 & \bf0.225 & 0.738 & 0.760
       & \bf0.009 & \bf0.318 & 0.780 & 0.797
       & \bf-1.321 & \bf-0.920 & 0.311 & 0.367 \\
     \bottomrule
  \end{tabular}
  \caption{Performance of six LMs under different risk levels for the third prompt variation.}
  \label{tab:main-prompt-3}
\end{table*}

\begin{table*}[t!]
  \tiny
  \centering
  \begin{tabular}{lcc|cc !{\vrule width 1.5pt} cc|cc !{\vrule width 1.5pt} cc|cc}
    \toprule
     \textbf{Dataset}
       & \multicolumn{4}{c}{\textbf{MedQA}}
       & \multicolumn{4}{c}{\textbf{MMLU}}
       & \multicolumn{4}{c}{\textbf{GPQA}} \\
     \midrule
     \textbf{Risk Level}
       & \multicolumn{2}{c}{\textbf{High-Risk}}
       & \multicolumn{2}{c}{\textbf{Low-Risk}}
       & \multicolumn{2}{c}{\textbf{High-Risk}}
       & \multicolumn{2}{c}{\textbf{Low-Risk}}
       & \multicolumn{2}{c}{\textbf{High-Risk}}
       & \multicolumn{2}{c}{\textbf{Low-Risk}} \\
     \textbf{$(p_{\mathrm{cor}}, p_{\mathrm{inc}})$}
       & (1,-8) & (1,-4)
       & (4,-1) & (8,-1)
       & (1,-8) & (1,-4)
       & (4,-1) & (8,-1)
       & (1,-8) & (1,-4)
       & (4,-1) & (8,-1) \\
     \toprule
     \multicolumn{13}{c}{\textbf{gpt-4o-2024-08-06}} \\
     \midrule
     No risk informed
       & 0.029 & 0.016 & 0.004 & 0.004
       & 0.024 & 0.013 & 0.003 & 0.003
       & 0.138 & 0.076 & 0.018 & 0.016 \\
     Risk-informing
       & 0.017 & 0.015 & 0.004 & 0.006
       & 0.030 & 0.028 & 0.004 & 0.001
       & 0.115 & 0.037 & 0.008 & 0.013 \\
     Stepwise prompt
       & 0.026 & 0.012 & 0.005 & 0.004
       & 0.041 & 0.012 & 0.010 & 0.008
       & 0.040 & 0.055 & 0.007 & 0.019 \\
     Prompt chaining
       & 0.017 & 0.027 & 0.003 & 0.003
       & 0.028 & 0.020 & 0.003 & 0.004
       & 0.030 & 0.086 & 0.019 & 0.016 \\
     \toprule
     \multicolumn{13}{c}{\textbf{gpt-4o-mini-2024-07-18}} \\
     \midrule
     No risk informed
       & 0.043 & 0.024 & 0.006 & 0.005
       & 0.035 & 0.019 & 0.005 & 0.004
       & 0.072 & 0.040 & 0.010 & 0.009 \\
     Risk-informing
       & 0.046 & 0.033 & 0.009 & 0.006
       & 0.081 & 0.032 & 0.007 & 0.011
       & 0.151 & 0.033 & 0.016 & 0.009 \\
     Stepwise prompt
       & 0.046 & 0.346 & 0.006 & 0.007
       & 0.024 & 0.023 & 0.004 & 0.006
       & 0.175 & 0.089 & 0.014 & 0.021 \\
     Prompt chaining
       & 0.054 & 0.032 & 0.003 & 0.002
       & 0.012 & 0.006 & 0.006 & 0.003
       & 0.049 & 0.036 & 0.011 & 0.008 \\
     \toprule
     \multicolumn{13}{c}{\textbf{claude-3-5-sonnet-20241022}} \\
     \midrule
     No risk informed
       & 0.054 & 0.030 & 0.007 & 0.007
       & 0.011 & 0.006 & 0.001 & 0.001
       & 0.062 & 0.034 & 0.009 & 0.008 \\
     Risk-informing
       & 0.046 & 0.015 & 0.006 & 0.005
       & 0.047 & 0.016 & 0.009 & 0.004
       & 0.312 & 0.175 & 0.004 & 0.014 \\
     Stepwise prompt
       & 0.006 & 0.024 & 0.007 & 0.007
       & 0.019 & 0.014 & 0.003 & 0.004
       & 0.446 & 0.598 & 0.033 & 0.016 \\
     Prompt chaining
       & 0.025 & 0.029 & 0.009 & 0.007
       & 0.008 & 0.007 & 0.000 & 0.000
       & 0.050 & 0.062 & 0.006 & 0.008 \\
     \toprule
     \multicolumn{13}{c}{\textbf{claude-3-5-haiku-20241022}} \\
     \midrule
     No risk informed
       & 0.023 & 0.013 & 0.003 & 0.003
       & 0.032 & 0.018 & 0.005 & 0.004
       & 0.031 & 0.017 & 0.004 & 0.004 \\
     Risk-informing
       & 0.112 & 0.028 & 0.008 & 0.015
       & 0.027 & 0.037 & 0.009 & 0.011
       & 0.031 & 0.132 & 0.008 & 0.005 \\
     Stepwise prompt
       & 0.012 & 0.014 & 0.009 & 0.002
       & 0.044 & 0.027 & 0.007 & 0.003
       & 0.080 & 0.084 & 0.016 & 0.003 \\
     Prompt chaining
       & 0.014 & 0.018 & 0.002 & 0.003
       & 0.022 & 0.011 & 0.006 & 0.000
       & 0.120 & 0.075 & 0.005 & 0.005 \\
     \toprule
     \multicolumn{13}{c}{\textbf{gemini-1.5-pro-002}} \\
     \midrule
     No risk informed
       & 0.008 & 0.003 & 0.002 & 0.002
       & 0.056 & 0.032 & 0.011 & 0.010
       & 0.111 & 0.061 & 0.014 & 0.013 \\
     Risk-informing
       & 0.089 & 0.030 & 0.009 & 0.012
       & 0.022 & 0.036 & 0.006 & 0.002
       & 0.083 & 0.161 & 0.021 & 0.014 \\
     Stepwise prompt
       & 0.016 & 0.023 & 0.009 & 0.005
       & 0.019 & 0.030 & 0.002 & 0.002
       & 0.239 & 0.084 & 0.005 & 0.009 \\
     Prompt chaining
       & 0.083 & 0.006 & 0.005 & 0.005
       & 0.100 & 0.054 & 0.009 & 0.011
       & 0.187 & 0.076 & 0.054 & 0.040 \\
     \toprule
     \multicolumn{13}{c}{\textbf{gemini-1.5-flash-002}} \\
     \midrule
     No risk informed
       & 0.064 & 0.036 & 0.010 & 0.009
       & 0.040 & 0.024 & 0.008 & 0.008
       & 0.103 & 0.062 & 0.025 & 0.024 \\
     Risk-informing
       & 0.025 & 0.037 & 0.011 & 0.007
       & 0.028 & 0.015 & 0.001 & 0.002
       & 0.024 & 0.109 & 0.005 & 0.015 \\
     Stepwise prompt
       & 0.046 & 0.008 & 0.005 & 0.003
       & 0.033 & 0.011 & 0.011 & 0.007
       & 0.059 & 0.058 & 0.014 & 0.016 \\
     Prompt chaining
       & 0.043 & 0.039 & 0.013 & 0.011
       & 0.030 & 0.028 & 0.010 & 0.009
       & 0.161 & 0.085 & 0.019 & 0.027 \\
     \toprule
     \multicolumn{13}{c}{\textbf{Averaged Across All Models}} \\
     \midrule
     \rowcolor{Gray}
     No risk informed
       & 0.037 & 0.020 & 0.005 & 0.005
       & 0.033 & 0.019 & 0.006 & 0.005
       & 0.086 & 0.048 & 0.013 & 0.012 \\
     \rowcolor{Gray}
     Risk-informing
       & 0.056 & 0.026 & 0.008 & 0.008
       & 0.039 & 0.027 & 0.006 & 0.005
       & 0.119 & 0.108 & 0.010 & 0.012 \\
     \rowcolor{Gray}
     Stepwise prompt
       & 0.025 & 0.071 & 0.007 & 0.005
       & 0.030 & 0.020 & 0.006 & 0.005
       & 0.173 & 0.161 & 0.015 & 0.014 \\
     \rowcolor{Gray}
     Prompt chaining
       & 0.039 & 0.025 & 0.008 & 0.006
       & 0.033 & 0.020 & 0.006 & 0.005
       & 0.100 & 0.070 & 0.019 & 0.018 \\
     \bottomrule
  \end{tabular}
  \caption{Standard deviations of three prompt variations, of which the performance is listed in Table~\ref{tab:main-prompt-1}, \ref{tab:main-prompt-2}, and \ref{tab:main-prompt-3}.}
  \label{tab:main-stddev}
\end{table*}

\begin{table*}[t!]
  \tiny
  \centering
  \begin{tabular}{lccc|c !{\vrule width 1.5pt} ccc|c !{\vrule width 1.5pt} ccc|c}
    \toprule
     \textbf{Dataset} & \multicolumn{4}{c}{\textbf{MedQA}} & \multicolumn{4}{c}{\textbf{MMLU}} & \multicolumn{4}{c}{\textbf{GPQA}} \\
     \midrule
     \textbf{Metric}  & ACC($\uparrow$) & ECE($\downarrow$) & EVR($\uparrow$) & $R(\uparrow)$ & ACC($\uparrow$) & ECE($\downarrow$) & EVR($\uparrow$) & $R(\uparrow)$ & ACC($\uparrow$) & ECE($\downarrow$) & EVR($\uparrow$) & $R(\uparrow)$ \\
     \toprule
     \multicolumn{13}{c}{\textbf{gpt-4o-2024-08-06}} \\
     \midrule
     Stepwise  &0.902    &0.159    &0.003    &0.166    &0.859    &\bf0.204 &0.009    &-0.169   &0.477    &0.343    &0.019    &-3.294   \\
     Chaining  &\bf0.905 &\bf0.150 &\bf0.924 &\bf0.475 &\bf0.875 &0.216    &\bf0.969 &\bf0.327 &\bf0.504 &\bf0.310 &\bf0.981 &\bf-0.054\\
     \toprule
     \multicolumn{13}{c}{\textbf{gpt-4o-mini-2024-07-18}} \\
     \midrule
     Stepwise 
       &0.795 &0.289 &0.000 &-0.835
       &0.810 &0.277 &0.000 &-0.683
       &\bf0.398 &0.411 &0.006 &-4.334\\
     Chaining 
       &\bf0.797 &\bf0.223 &\bf0.015 &\bf-0.083
       &\bf0.814 &\bf0.195 &\bf0.033 &\bf-0.016
       &0.395 &\bf0.302 &\bf0.103 &\bf-0.969\\
     \toprule
     \multicolumn{13}{c}{\textbf{claude-3-5-sonnet-20241022}} \\
     \midrule
     Stepwise 
       &0.885 &0.182 &0.260 &0.083
       &0.881 &\bf0.147 &0.642 &0.197
       &0.542 &0.304 &0.504 &-1.756\\
     Chaining 
       &\bf0.894 &\bf0.165 &\bf0.974 &\bf0.324
       &\bf0.887 &0.186 &\bf0.984 &\bf0.307
       &\bf0.568 &\bf0.263 &\bf0.979 &\bf-0.413\\
     \toprule
     \multicolumn{13}{c}{\textbf{claude-3-5-haiku-20241022}} \\
     \midrule
     Stepwise 
       &\bf0.772 &0.194 &0.000 &-1.034
       &0.796 &0.190 &0.001 &-0.762
       &\bf0.380 &0.397 &0.002 &-4.502\\
     Chaining 
       &0.771 &\bf0.182 &\bf0.041 &\bf-0.835
       &\bf0.808 &\bf0.117 &\bf0.247 &\bf-0.305
       &0.367 &\bf0.392 &\bf0.163 &\bf-3.129\\
     \toprule
     \multicolumn{13}{c}{\textbf{gemini-1.5-pro-002}} \\
     \midrule
     Stepwise 
       &\bf0.822 &\bf0.202 &0.000 &-0.565
       &\bf0.850 &\bf0.118 &0.001 &-0.273
       &0.541 &0.340 &0.000 &-2.901\\
     Chaining 
       &0.814 &0.482 &\bf0.447 &\bf-0.316
       &0.848 &0.213 &\bf0.693 &\bf-0.066
       &\bf0.555 &\bf0.297 &\bf0.589 &\bf-2.127\\
     \toprule
     \multicolumn{13}{c}{\textbf{gemini-1.5-flash-002}} \\
     \midrule
     Stepwise 
       &0.727 &\bf0.256 &0.000 &-1.414
       &0.811 &0.267 &0.000 &-0.648
       &0.437 &\bf0.255 &0.004 &-3.815\\
     Chaining 
       &\bf0.733 &0.438 &\bf0.099 &\bf-0.537
       &\bf0.813 &\bf0.256 &\bf0.322 &\bf-0.175
       &\bf0.461 &0.266 &\bf0.193 &\bf-1.178\\
     \toprule
     \multicolumn{13}{c}{\textbf{Averaged Across All Models}} \\
     \midrule
     \rowcolor{Gray}
     Stepwise  
       &0.817 &\bf0.214 &0.044 &-0.600
       &0.835 &0.201 &0.109 &-0.389
       &0.462 &0.341 &0.089 &-3.434\\
     \rowcolor{Gray}
     Chaining 
       &\bf0.819 &0.273 &\bf0.417 &\bf-0.162
       &\bf0.841 &\bf0.197 &\bf0.541 &\bf0.012
       &\bf0.475 &\bf0.305 &\bf0.501 &\bf-1.312\\
     \bottomrule
  \end{tabular}
  \caption{Performance of six LMs for different performance metrics. All scores are rounded to three decimal places. Abbreviations: Accuracy, ACC; Expected Calibration Error, ECE; proportions of Expected-Value Reasoning, EVR. Looking closely, prompt chaining performs better than stepwise prompts in 13 out of 18 settings (3 datasets $\times$ 6 LMs = 18) at ACC, 12 out of 18 settings at ECE, and 18 out of 18 at EVR.}
  \label{tab:stepwise-vs-chaining}
\end{table*}

\end{document}